\documentclass{article} % For LaTeX2e
\PassOptionsToPackage{dvipsnames}{xcolor}

\usepackage[utf8]{inputenc} % allow utf-8 input
\usepackage[T1]{fontenc}    % use 8-bit T1 fonts
\usepackage[
colorlinks=true,
linkcolor=black,
filecolor=black,
urlcolor=CadetBlue,
citecolor=CadetBlue
]{hyperref}

\newif\ifpreprint
\preprinttrue

\ifpreprint
  \usepackage[preprint]{icml2026}
  \hypersetup{pdfsubject={}}
\else
  \usepackage[accepted]{icml2026}
\fi

\usepackage{url}            % simple URL typesetting
\usepackage{xurl}           % allow URL line breaks at any character
\providecommand{\doi}[1]{\href{https://doi.org/#1}{doi:\nolinkurl{#1}}}
\usepackage{booktabs}       % professional-quality tables
\usepackage{nicefrac}       % compact symbols for 1/2, etc.
\usepackage{microtype}      % microtypography
\usepackage{graphicx}

\usepackage{amsmath,amssymb,amsfonts}
\usepackage{bbm}
\usepackage{bm}
\usepackage{mathtools}
\usepackage[automake]{glossaries}
\usepackage{xspace,mfirstuc,tabulary}
\usepackage{multirow}
\usepackage{balance}
\usepackage{listings}
\usepackage{subfig}
\usepackage[subfigure]{tocloft}
\usepackage{makecell}
\usepackage{float}
\usepackage{etoolbox}
\usepackage{seqsplit}
\usepackage{pgfplots}
\usepackage{enumitem}
\setlist[itemize]{itemindent=0.1em, leftmargin=1.5em, itemsep=1pt, topsep=3pt}
\setlist[enumerate]{itemindent=0.1em, leftmargin=1.5em, itemsep=1pt, topsep=3pt}
\usepackage[capitalize,noabbrev]{cleveref}

\pgfplotsset{compat=1.18}
\usetikzlibrary{calc,arrows.meta,fit}

\definecolor{annotcolor}{RGB}{25,60,150}  % annotator traits (navy)
\definecolor{itemcolor}{RGB}{150,25,35}   % item traits (oxblood)
\definecolor{labelcolor}{RGB}{20,105,55}  % labels (bottle green)
\definecolor{clfcolor}{RGB}{165,85,5}     % classifier (umber)
\definecolor{noisecolor}{RGB}{0,110,118}  % noise model (dark teal)
\DeclareRobustCommand{\cann}[1]{{\color{annotcolor}#1}}
\DeclareRobustCommand{\citm}[1]{{\color{itemcolor}#1}}
\DeclareRobustCommand{\clab}[1]{{\color{labelcolor}#1}}
\DeclareRobustCommand{\cclf}[1]{{\color{clfcolor}#1}}
\DeclareRobustCommand{\cnoise}[1]{{\color{noisecolor}#1}}

\icmltitlerunning{Learning from Crowds with Unsupervised Item Response Models}

\newacronym{IRT}{IRT}{Item Response Theory}
\newacronym{SVI}{SVI}{Stochastic Variational Inference}
\newacronym{MCMC}{MCMC}{Markov Chain Monte Carlo}
\newacronym{ELBO}{ELBO}{Evidence Lower Bound}
\newacronym{ECE}{ECE}{Expected Calibration Error}
\newacronym{MML}{MML}{Marginal Maximum Likelihood}
\newacronym{MIRT}{MIRT}{Multidimensional Item Response Theory}
\newacronym{GLAD}{GLAD}{Generative model of Labels, Abilities, and Difficulties}
\newacronym{NLP}{NLP}{Natural Language Processing}
\newacronym{ML}{ML}{Machine Learning}
\newacronym{ICC}{ICC}{Item Characteristic Curve}
\newacronym{MLP}{MLP}{Multilayer Perceptron}
\newacronym{EM}{EM}{Expectation Maximisation}
\newacronym{LLM}{LLM}{Large Language Model}
\newacronym{MV}{MV}{majority vote}
\newacronym{MFCC}{MFCC}{Mel-Frequency Cepstral Coefficients}
\newacronym{MAP}{MAP}{Maximum A Posteriori}
\newacronym{RTE}{RTE}{Recognising Textual Entailment}
\newacronym{MNLI}{MNLI}{Multi-Genre Natural Language Inference}
\newacronym{AMT}{AMT}{Amazon Mechanical Turk}

\makeglossaries
\begin{document}

\twocolumn[
\icmltitle{Learning from Crowds with Unsupervised Item Response Models}

\begin{icmlauthorlist}
\icmlauthor{Pascal Wullschleger}{mu,hslu}
\icmlauthor{Majid Zarharan}{dcu}
\icmlauthor{Donnacha Daly}{hslu}
\icmlauthor{Marc Pouly}{hslu}
\icmlauthor{Jennifer Foster}{mu}
\end{icmlauthorlist}

\icmlaffiliation{mu}{Hamilton Institute, Maynooth University, Ireland}
\icmlaffiliation{hslu}{Lucerne School of Computer Science and IT, Switzerland}
\icmlaffiliation{dcu}{School of Computing, Dublin City University, Ireland}

\icmlcorrespondingauthor{Pascal Wullschleger}{pascal.wullschleger@hslu.ch}

\icmlkeywords{Item Response Theory, Crowdsourcing, Learning from Crowds, Label Inference}

\vskip 0.3in
]

\printAffiliationsAndNotice{}

\begin{abstract}
\gls{IRT} is the study of tests, including the properties of both test items
and test takers.
Crowdsourced labelling can be cast as an unsupervised \gls{IRT} problem:
annotators play the role of test takers with varying ability,
and annotation tasks play the role of items with varying
difficulty. We evaluate unsupervised variants of standard
logistic \gls{IRT} models for latent trait recovery and label inference on
six datasets with crowdsourced labels and known ground truths.
Unsupervised \gls{IRT} models can recover annotator and item traits
(e.g., ability, difficulty and discrimination) reliably across most datasets
when compared to supervised variants trained on the ground truth.
\gls{IRT} models seem to be more stable in label inference compared to most
baselines, but might not be the best possible choice depending on the dataset.
We further propose to use \gls{IRT} models to inform classifiers during
training and compare against existing crowd-learning methods.
We find that crowd-learning methods can improve F1 over majority-vote
baselines, but no method significantly improves both F1 and calibration.
\gls{IRT}-informed curriculum methods seem to preserve calibration better
than other crowd-learning methods, but are not at the frontier regarding F1, and the best
crowd-learning methods in terms of F1 seem to be miscalibrated.\textsuperscript{*}
\end{abstract}
\glsresetall

\section{Introduction}
Crowdsourcing is a common approach to annotating datasets in order to collect 
labels where they were previously unavailable. 
Recovering true labels from noisy annotations from annotators with different skills
and reliabilities is called label inference \citep{10.14778/3055540.3055547}.
Some work suggests that crowdsourced labels contain signal relevant
for model training beyond an aggregation of the annotations as hard or soft labels \citep{10.1613/jair.1.12752,plank-2022-problem}.
For example, without knowledge about the individual annotator behaviour, we cannot decide on their reliability regarding a task.
Indeed, there are methods to explicitly train classifiers on the raw annotations without aggregation, 
such as \citep{Rodrigues_Pereira_2018,Chu_Ma_Wang_2021,Herde2024} and many more.

\begin{figure*}[t]
	\centering
    \subfloat[Difficulty $\citm{\beta}_i$\label{fig:icc:beta}]{%
		\begin{tikzpicture}
			\begin{axis}[
					width=0.30\textwidth,
					height=0.24\textwidth,
					xlabel={\scriptsize $\cann{\theta}_j$},					
                    ylabel={\scriptsize $P_{\bm{\psi}}(\clab{L}_{ij} = \clab{Z}_i)$},
					xmin=-4, xmax=4, ymin=0, ymax=1,
					domain=-4:4, samples=100,
					legend style={at={(0.97,0.03)}, anchor=south east,
							font=\scriptsize, row sep=-2pt, inner sep=1pt,
							nodes={scale=0.65},
							/tikz/every even column/.append style={column sep=0pt}},
					legend image post style={scale=0.5},
					tick label style={font=\scriptsize},
					grid=major, grid style={gray!30},
				]
				\addplot[blue, thick, dotted] {1/(1+exp(-1*(x-(-1))))};
				\addplot[black, thick]         {1/(1+exp(-1*(x-0)))};
				\addplot[red, thick, dashed]   {1/(1+exp(-1*(x-1)))};
				\legend{$\citm{\beta}_i\!=\!-1.0$, $\citm{\beta}_i\!=\!0.0$,
					$\citm{\beta}_i\!=\!1.0$}
			\end{axis}
		\end{tikzpicture}%
	}%
	\subfloat[Discrimination $\citm{\alpha}_i$\label{fig:icc:alpha}]{%
		\begin{tikzpicture}
			\begin{axis}[
					width=0.30\textwidth,
					height=0.24\textwidth,
					xlabel={\scriptsize $\cann{\theta}_j$},
					yticklabels={},
					xmin=-4, xmax=4, ymin=0, ymax=1,
					domain=-4:4, samples=100,
					legend style={at={(0.97,0.03)}, anchor=south east,
							font=\scriptsize, row sep=-2pt, inner sep=1pt,
							nodes={scale=0.65},
							/tikz/every even column/.append style={column sep=0pt}},
					legend image post style={scale=0.5},
					tick label style={font=\scriptsize},
					grid=major, grid style={gray!30},
				]
				\addplot[blue, thick, dotted] {1/(1+exp(-0.5*(x-0)))};
				\addplot[black, thick]         {1/(1+exp(-1*(x-0)))};
				\addplot[red, thick, dashed]   {1/(1+exp(-2*(x-0)))};
				\legend{$\citm{\alpha}_i\!=\!0.5$, $\citm{\alpha}_i\!=\!1.0$,
					$\citm{\alpha}_i\!=\!2.0$}
			\end{axis}
		\end{tikzpicture}%
	}%
	\subfloat[Guessing $\citm{c}_i$\label{fig:icc:guess}]{%
		\begin{tikzpicture}
			\begin{axis}[
					width=0.30\textwidth,
					height=0.24\textwidth,
					xlabel={\scriptsize $\cann{\theta}_j$},
					yticklabels={},
					xmin=-4, xmax=4, ymin=0, ymax=1,
					domain=-4:4, samples=100,
					legend style={at={(0.97,0.03)}, anchor=south east,
							font=\scriptsize, row sep=-2pt, inner sep=1pt,
							nodes={scale=0.65},
							/tikz/every even column/.append style={column sep=0pt}},
					legend image post style={scale=0.5},
					tick label style={font=\scriptsize},
					grid=major, grid style={gray!30},
				]
				\addplot[black, thick]
				{0 + (1-0)*1/(1+exp(-1*(x-0)))};
				\addplot[blue, thick, dotted]
				{0.25 + (1-0.25)*1/(1+exp(-1*(x-0)))};
				\addplot[red, thick, dashed]
				{0.5 + (1-0.5)*1/(1+exp(-1*(x-0)))};
				\legend{$\citm{c}_i\!=\!0.00$, $\citm{c}_i\!=\!0.25$, $\citm{c}_i\!=\!0.50$}
			\end{axis}
		\end{tikzpicture}%
	}%
	\subfloat[GLAD difficulty $1/\citm{\gamma}_i$\label{fig:icc:glad}]{%
		\begin{tikzpicture}
			\begin{axis}[
					width=0.30\textwidth,
					height=0.24\textwidth,
					xlabel={\scriptsize $\cann{\theta}_j$},
					yticklabels={},
					xmin=-4, xmax=4, ymin=0, ymax=1,
					domain=-4:4, samples=100,
					legend style={at={(0.97,0.03)}, anchor=south east,
							font=\scriptsize, row sep=-2pt, inner sep=1pt,
							nodes={scale=0.65},
							/tikz/every even column/.append style={column sep=0pt}},
					legend image post style={scale=0.5},
					tick label style={font=\scriptsize},
					grid=major, grid style={gray!30},
				]
				\addplot[blue, thick, dotted]     {1/(1+exp(-exp(0.5)*(x)))};
				\addplot[black, thick]             {1/(1+exp(-exp(1.0)*(x)))};
				\addplot[red, thick, dashed]       {1/(1+exp(-exp(2.0)*(x)))};
				\legend{$1/\citm{\gamma}_i\!=\!2.0$, $1/\citm{\gamma}_i\!=\!1.0$,
					$1/\citm{\gamma}_i\!=\!0.5$}
			\end{axis}
		\end{tikzpicture}%
	}
	\caption{\glspl{ICC} with varying latent trait values.
		(a--c) show the effect of each 3PL parameter:
		difficulty $\citm{\beta}_i$ shifts the curve horizontally,
		discrimination $\citm{\alpha}_i$ controls slope steepness, and
		guessing $\citm{c}_i$ sets the lower asymptote.
		(d) shows the GLAD model: $e^{\citm{\gamma}_i}$ scales the slope
		and the inflection point is always fixed at $\cann{\theta}_j = 0$;
		curves are labelled by GLAD-difficulty $1/\citm{\gamma}_i$.}
	\label{fig:icc}
\end{figure*}

\acrfull{IRT} is the modern study of tests, which is used to compare the abilities of test takers, infer item properties 
such as difficulty, or even dynamically adapt test questions during test administration \citep{Fox2010-nd}. 
This field appears closely related to crowdsourcing when we assume that
the quality of labels heavily depends on the ability of the annotators and
the difficulty of the examples.
Research on \gls{IRT} dates back to at least the 1960s
\citep{rasch1960, birnbaum1968}. 

While a breadth of label inference approaches has been compared to standard \gls{MV}, from models fitting confusion matrices 
of annotators \citep{dawid_skene}, leveraging correlations between annotators \citep{pmlr-v97-li19i} to \gls{IRT}-like methods \citep{glad}, the
application of classical interpretable logistic \gls{IRT} models to label inference remains under-explored.

Commonly, \gls{IRT} models assume the true labels to be known and the focus is on extracting 
latent traits of items and annotators, such as difficulty and ability.
While some work has studied \gls{IRT} for label inference when
ground truths are partly available \citep{khattak_irt}, the application of 
\gls{IRT} in an unsupervised setting has seen limited interest. 

We argue that label inference is a natural 
application of unsupervised \gls{IRT}, since it is like taking a test to which the correct answers are presently unknown.
In this work, we investigate whether logistic \gls{IRT} models can be
used in an unsupervised manner for label inference, thus
bridging the gap to well-established methodologies in test theory and
recovering interpretable latent traits of items and
annotators.
We also explore whether the latent traits found by \gls{IRT} models can be used to inform downstream model training.%
{\renewcommand{\thefootnote}{*}\footnote{Code will be released upon peer-reviewed publication.}\addtocounter{footnote}{-1}}

We address the following research questions in this work:
\begin{itemize}
	\item \textbf{RQ1:} Can we recover the latent traits of items and annotators in an unsupervised setting?
    \item \textbf{RQ2:} Are standard \gls{IRT} models useful for label inference from crowdsourced annotations?
    \item \textbf{RQ3:} Can we incorporate latent traits found by \gls{IRT} models as signal to improve or calibrate models trained on crowdsourced data?
\end{itemize}

We conduct experiments on six datasets containing expert ground truths and
normal annotator votes and find that latent traits recovered by unsupervised
models are highly correlated with traits inferred by supervised models on most
datasets. Trait recovery also seems to become more difficult as the number of
traits increases. \gls{IRT} models are competitive on label inference across
the different datasets and seem to be more reliable overall than baselines.
The estimated annotator and item traits can be used to improve downstream
classifier performance by forward correction compared to baselines uninformed
by crowd-labels, but do not reach state-of-the-art performance overall.
However, \gls{IRT}-informed methods seem to preserve calibration better than
other crowd-learning classifiers.

\section{Item Response Theory}

\gls{IRT} provides a framework for modelling the
relationship between latent traits, such as the difficulty of items
and the abilities of respondents, and observed responses
\citep{Fox2010-nd}. In the following sections, we introduce the most common logistic models and their respective assumptions.
We write $P_{\bm{\psi}}$ for probabilities defined by a model with latent
traits $\bm{\psi}$.

\paragraph{One-Parameter Logistic (1PL) Model.}

The 1PL model, also known as the Rasch model \citep{rasch1960}, assumes
that the probability of a correct response depends solely on the
difference between annotator ability and item difficulty
(Eq.~\ref{eq:1pl}). The curve formed by the logistic model for a single
item is called the 
\acrfull{ICC}
(Fig.~\ref{fig:icc}).

\begin{equation}
	P_{\bm{\psi}}(\clab{L}_{ij} = \clab{Z}_i) = \sigma(\cann{\theta}_j - \citm{\beta}_i)
	\label{eq:1pl}
\end{equation}

Here, $\clab{L}_{ij}$ is the response of annotator $j$ to item $i$,
$\clab{Z}_i$ denotes the correct answer to item $i$,
$\cann{\theta}_j \in \mathbb{R}$ represents the ability of annotator $j$,
$\citm{\beta}_i \in \mathbb{R}$ represents the difficulty of item $i$, and
$\sigma(x) = \frac{1}{1 + e^{-x}}$ is the logistic function.
The latent traits are $\bm{\psi} = (\cann{\bm{\theta}}, \citm{\bm{\beta}})$.
Higher-ability annotators have a greater probability of correctly
responding to items of constant difficulty, while more difficult items
uniformly reduce this probability across equally skilled annotators
(see Fig.~\ref{fig:icc:beta}).

\paragraph{Two-Parameter Logistic (2PL) Model.}

The 2PL model (Eq.~\ref{eq:2pl}) as introduced by \citet{birnbaum1968}
extends the 1PL by introducing item-specific discrimination parameters
$\citm{\alpha}_i$.

\begin{equation}
	P_{\bm{\psi}}(\clab{L}_{ij} = \clab{Z}_i) =
	    \sigma(\citm{\alpha}_i(\cann{\theta}_j - \citm{\beta}_i))
	\label{eq:2pl}
\end{equation}

where $\citm{\alpha}_i \in \mathbb{R}^+$ represents the discrimination of item
$i$, while $\cann{\theta}_j$ and $\citm{\beta}_i$ are defined as in the 1PL model,
and $\bm{\psi} = (\cann{\bm{\theta}}, \citm{\bm{\alpha}}, \citm{\bm{\beta}})$.
The discrimination parameter $\citm{\alpha}_i$ controls how steeply the response
probability changes with ability.
Discriminative items more effectively differentiate between annotators
of different abilities (e.g., does a test question reward a certain skill)
in that the slope of the \gls{ICC} is
steeper and the tails thus converge faster (see Fig.~\ref{fig:icc:alpha}). 

\paragraph{Three-Parameter Logistic (3PL) Model.}

The 3PL model (Eq.~\ref{eq:3pl}), as introduced by \citet{birnbaum1968},
adds a guessing parameter $\citm{c}_i \in [0, 1)$ that sets a lower asymptote
on the item response function, representing the probability that an
annotator with very low ability will still provide a correct response
simply by guessing.
This is particularly relevant in settings where random guesses result
in non-zero accuracy (e.g.,\ multiple-choice, see Fig.~\ref{fig:icc:guess}).
The latent traits are
$\bm{\psi} = (\cann{\bm{\theta}}, \citm{\bm{\alpha}}, \citm{\bm{\beta}}, \citm{\bm{c}})$.

\begin{equation}
	\label{eq:3pl}
	P_{\bm{\psi}}(\clab{L}_{ij} = \clab{Z}_i) =
	    \citm{c}_i + (1 - \citm{c}_i)\,\sigma(\citm{\alpha}_i(\cann{\theta}_j - \citm{\beta}_i))
\end{equation}

\paragraph{Trait Estimation.}
\gls{IRT} models typically estimate traits via \gls{MML} and the \gls{EM} algorithm, which
integrates out latent abilities \citep{Fox2010-nd}.
Bayesian approaches specify prior distributions and use \gls{MCMC}
or \gls{SVI} for posterior approximation.
\gls{SVI} mainly offers better computational efficiency for large-scale
applications while maintaining principled uncertainty quantification  \citep{10.3389/fpsyg.2016.01422}. 

\section{Related Work}

\subsection{Difficulty of Training Examples in ML}

The difficulty of training examples influences \gls{ML} models in several ways and has been studied and leveraged for training through a wide range of approaches.
Curriculum learning, for instance, can be used to vary example difficulty throughout training and can improve the performance of ML models. 
However, such approaches can be difficult to use since they require a method for ordering samples from easiest to most difficult, as well as a proper schedule for when to add more difficult data during training \citep{soviany2022curriculum}. 
\citet{baldock2021deep} report that early layers generalise while later layers memorise, and networks learn easy data and simpler functions first.
\citet{lee2025regularizing} look into the negative effect of difficult examples on generalisation, especially in adversarial training. In theory, they show that when adversarial training is used, difficult examples become even more difficult. 
\citet{bayat2025pitfalls} explore how memorisation and generalisation affect each other. When a model memorises spurious correlations, it often reduces generalisation. 

\subsection{Applications of \gls{IRT} in \gls{ML}} 
\label{sec:dds}

\citet{MARTINEZPLUMED201918} apply \gls{IRT} to classification problems 
in order to explore how \gls{IRT} parameters, like discrimination, difficulty, and guessing, 
relate to classification instances and how they connect with measures of instance difficulty.
They also look into how the estimated ability of a classifier could be used to compare classifier performance. 
\citet{lalor-yu-2020-dynamic} suggest a new approach: 
instead of using heuristic methods to estimate difficulty,  difficulty parameters are learned through \gls{IRT}.
At each training epoch, the method checks the model's current ability level and 
picks the most suitable examples to train on. Models that use these learned difficulty and ability measures perform better than those that rely 
on heuristic-based curriculum learning when tested on GLUE \citep{glue} classification tasks.

\citet{PUDF} extend this approach into a framework that builds the
artificial crowd from multiple pre-trained models
instead of a single model type (LSTM or BERT).
They also apply the dynamic data selection to larger models.

\gls{IRT} has also been applied to \gls{NLP} leaderboards to mitigate the fact that not all test examples are equally informative.
Thus, the same raw test score can have different implications depending on which items were correctly answered \citep{rodriguez2021evaluation}.

\subsection{Label Inference with \gls{IRT}}

\paragraph{Semi-Supervised Methods.}
\citet{khattak_irt} perform label inference by fitting \gls{IRT} models on
a subset of expert-labelled instances and subsequently reusing
parameters on the remaining samples to aggregate non-expert labels by
marginalising out the annotator abilities.

\paragraph{Unsupervised Methods.}
Perhaps the most closely related work to unsupervised \gls{IRT} to date is the GLAD model
\citep{glad}, which jointly infers annotator abilities
$\cann{\theta}_j$, item difficulties $1/\citm{\gamma}_i$, and true labels $\clab{Z}_i$.
The model specifies the probability that annotator $j$ correctly labels
item $i$ as shown in Eq.~\ref{eq:glad}.

\begin{equation}
	P_{\bm{\psi}}(\clab{L}_{ij} = \clab{Z}_i) = \sigma(\cann{\theta}_j e^{\citm{\gamma}_i})
	\label{eq:glad}
\end{equation}

We use $\citm{\gamma}_i$ rather than $\citm{\beta}_i$ to avoid confusion with the
standard \gls{IRT} difficulty parameter: in GLAD, $1/\citm{\gamma}_i$ plays the role of
difficulty, so $\citm{\gamma}_i$ is better understood as an inverse-difficulty.
In the unsupervised case the true label $\clab{Z}_i$ is unknown,
and its estimate $\clab{\hat{z}}_i$ is jointly inferred with the other
parameters.
This formulation treats $e^{\citm{\gamma}_i}$ as a multiplicative scaling
factor rather than a shift, so the inflection point is always fixed at
$\cann{\theta}_j = 0$ regardless of item difficulty (see
Fig.~\ref{fig:icc:glad}).
However, \gls{IRT} has long-established interpretations of these
traits.
For instance, the slope of the curve is generally interpreted as
discrimination, while ability and difficulty represent shifts \citep{Fox2010-nd}.
In GLAD, both $\cann{\theta}_j$ and $e^{\citm{\gamma}_i}$ only affect the slope
due to the product.
We argue that the GLAD-style inference should also be
studied using standard \gls{IRT} models to preserve parameter interpretations.

\citet{mace} introduce an unsupervised
item response model for label inference, where the ability represents trustworthiness, 
i.e.\ the probability of an annotator not spamming on a task. 
However, they are not fitting a standard logistic \gls{IRT} model, but rather
a mixture model with a different interpretation.
We use both of these models as baselines
in our experiments and compare them to the standard logistic \gls{IRT} models. 

\subsection{Learning from Crowds}
Various methods have been proposed to learn from crowdsourced labels directly, without
requiring inference of a ground truth beforehand \citep{Rodrigues_Pereira_2018,Chu_Ma_Wang_2021,Herde2024,Herde2025crowdhpoRH}. 
A challenge in learning from crowds is that many approaches estimate
annotator confusions or noise models without incorporating feature
information directly.
\citet{lfc} argue that the features should be used to inform the inferred
label and the annotator models jointly while training classifiers.
Perhaps the most common approach to this is to corrupt model predictions with a model of annotator noise, before calculating
a loss against the observed annotator labels \citep{Chu_Ma_Wang_2021,Rodrigues_Pereira_2018,Herde2024}. 
This is called forward correction, and has been proposed
using a variety of
noise models, such as simple confusion matrices per-annotator \citep{Rodrigues_Pereira_2018},
or a combination of common confusions and per-annotator residuals \citep{Chu_Ma_Wang_2021}.
AnnotMix \citep{Herde2024} extends this paradigm with mixup regularisation \citep{mixup},
which assumes that interpolation of feature vectors and annotators should translate to interpolations of the targets.

\section{Methods}
In this section we first introduce our method inspired by \citet{glad} for fitting
\gls{IRT} models in an unsupervised mode by simultaneously inferring labels with latent traits.
Using this method, we can later evaluate how well latent item and annotator traits are recovered (RQ1) and
how accurate inferred true labels are (RQ2). Subsequently, we propose methods to leverage the latent traits
in classifier training through forward correction and curriculum learning (RQ3).

\subsection{Unsupervised Trait Inference}
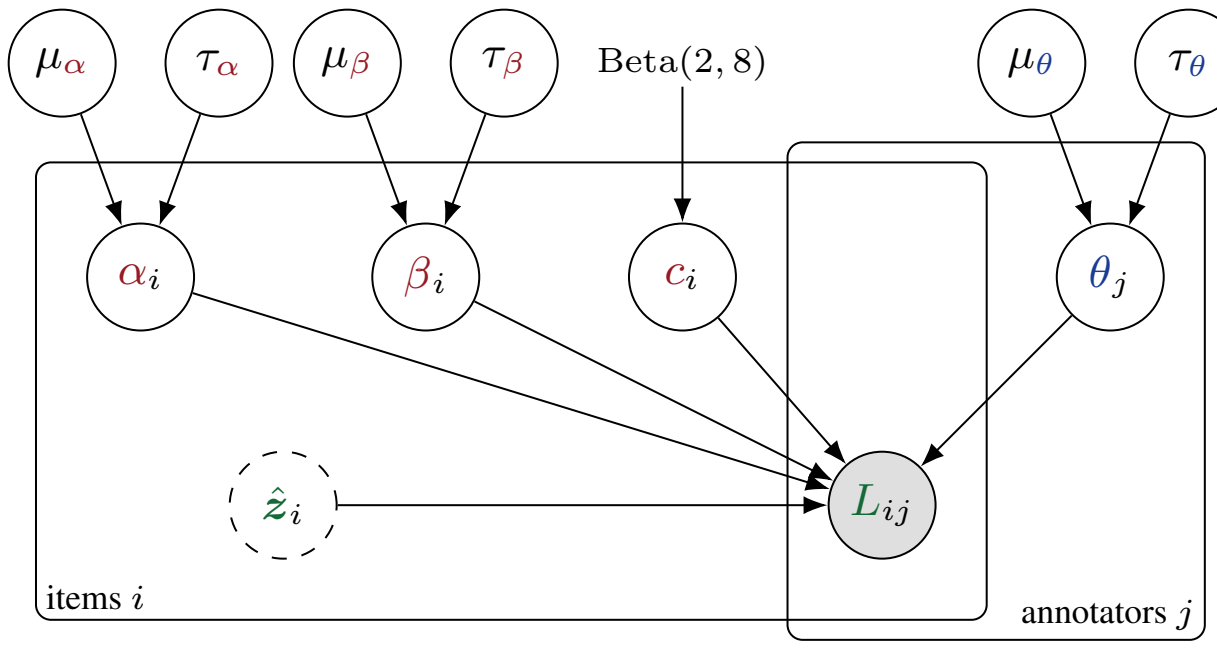
\begin{figure}[t]
  \centering
  \resizebox{\linewidth}{!}{% Bayesian network of the hierarchical 3PL crowdsourcing IRT model.
% Requires: \usetikzlibrary{arrows.meta,fit}

\begin{tikzpicture}[
  font=\small,
  latent/.style={
    circle, draw, thin, minimum size=0.75cm, inner sep=0pt},
  obs/.style={latent, fill=gray!25},
  est/.style={latent, dashed},
  const/.style={
    rectangle, inner sep=1pt, font=\scriptsize},
  plate/.style={
    draw, rounded corners=3pt, thin, inner sep=6pt},
  platelabel/.style={
    font=\scriptsize, anchor=south east, inner sep=2pt},
  arr/.style={->, >=Latex, thin},
]

%--- Hyperpriors ---
\node[latent] (mua) at (-0.55, 3.3) {$\mu_{\citm{\alpha}}$};
\node[latent] (taua) at ( 0.55, 3.3) {$\tau_{\citm{\alpha}}$};
\node[latent] (mub) at ( 1.45, 3.3) {$\mu_{\citm{\beta}}$};
\node[latent] (taub) at ( 2.55, 3.3) {$\tau_{\citm{\beta}}$};
\node[const] (cprior) at ( 3.8, 3.3) {$\mathrm{Beta}(2, 8)$};
\node[latent] (mut) at ( 6.25, 3.3) {$\mu_{\cann{\theta}}$};
\node[latent] (taut) at ( 7.35, 3.3) {$\tau_{\cann{\theta}}$};

%--- Item and annotator traits ---
\node[latent] (alpha) at ( 0.0, 1.8) {$\citm{\alpha}_i$};
\node[latent] (beta) at ( 2.0, 1.8) {$\citm{\beta}_i$};
\node[latent] (c) at ( 3.8, 1.8) {$\citm{c}_i$};
\node[latent] (theta) at ( 6.8, 1.8) {$\cann{\theta}_j$};

%--- Label estimate and observed annotation ---
\node[est] (z) at ( 1.0, 0.2) {$\clab{\hat{\bm{z}}}_i$};
\node[obs] (L) at ( 5.2, 0.2) {$\clab{L}_{ij}$};

%--- Edges ---
\draw[arr] (mua) -- (alpha);
\draw[arr] (taua) -- (alpha);
\draw[arr] (mub) -- (beta);
\draw[arr] (taub) -- (beta);
\draw[arr] (cprior) -- (c);
\draw[arr] (mut) -- (theta);
\draw[arr] (taut) -- (theta);

\draw[arr] (alpha) -- (L);
\draw[arr] (beta) -- (L);
\draw[arr] (c) -- (L);
\draw[arr] (theta) -- (L);
\draw[arr] (z) -- (L);

%--- Plates ---
\node[plate, fit=(alpha)(c)(z)(L), inner xsep=10pt,
  inner ysep=12pt] (items) {};
\node[platelabel, anchor=south west] at (items.south west) {items $i$};
\node[plate, fit=(theta)(L), inner xsep=8pt,
  inner ysep=16pt] (annots) {};
\node[platelabel] at (annots.south east) {annotators $j$};

\end{tikzpicture}}
  \caption{The Bayesian network for the unsupervised 3PL model. $\clab{L}_{ij}$ is the observed label on item $i$ given
  by annotator $j$. The \gls{IRT}-traits are latent and $\clab{\hat{\bm{z}}}_i$ is not a random variable, since it is set in a separate label step 
with fixed latents. The different $\mu$ and $\tau$ parameterise the learned Gaussian priors of the model.}
  \label{fig:irt_bayesian_network}
\end{figure}

In the following sections, we introduce our method using the example of
the 3PL model. By setting the guessing parameter $\citm{c}_i = 0$, the 3PL model
reduces to the 2PL model. Further setting the discrimination
parameter $\citm{\alpha}_i = 1$ yields the 1PL model.
We adopt the formulation proposed by \citet{glad} in GLAD
to model the generative probability of a correct annotation as
$p_{ij} \coloneqq P_{\bm{\psi}}(\clab{L}_{ij} = \clab{Z}_i)$
(Eq.~\ref{eq:3pl}),
where $\clab{Z}_i$ is the unobserved true label of item $i$.
As in GLAD, $\clab{Z}_i$ is handled in an \gls{EM} fashion:
we keep a soft label estimate $\clab{\hat{\bm{z}}}_i$, the posterior probability
of each class, which is recomputed in closed form rather than learned.
However, in contrast to GLAD, which point estimates the latent traits
in the M-step, we approximate the posterior over the traits,
$p(\bm{\psi} \mid \clab{\bm{L}}, \clab{\hat{\bm{z}}})$,
with a variational distribution
$q(\bm{\psi})$.
The resulting variational-EM hybrid alternates two steps,
starting from $\clab{\hat{\bm{z}}}$ set to the majority vote.
In the \emph{label step}, $\clab{\hat{\bm{z}}}_i$ is set to the closed-form posterior
$P_{\bar{\bm{\psi}}}(\clab{Z}_i \mid \clab{\bm{L}}_i)$ under a uniform class prior, with the traits
fixed at point estimates $\bar{\bm{\psi}}$ from $q$.
In the \emph{trait step}, $\clab{\hat{\bm{z}}}$ is held fixed and $q$ is
updated with one step of \gls{SVI}. The Bayesian network for 
the trait step is depicted in Fig.~\ref{fig:irt_bayesian_network}.

\paragraph{Priors.}
We follow \citet{lalor2019emnlp} and use hierarchical priors for ability ($\cann{\theta}_j$), difficulty
($\citm{\beta}_i$), and discrimination ($\citm{\alpha}_i$).
Each group has independent hyperpriors
$\mu \sim \mathcal{N}(0, 1)$ and $\tau \sim \text{HalfNormal}(1)$,
with $\cann{\theta}_j, \citm{\beta}_i \sim \mathcal{N}(\mu, \tau)$ and
$\citm{\alpha}_i \sim \text{LogNormal}(\mu, \tau)$.
A beta prior constrains guessing to $[0, 1]$: $\citm{c}_i \sim \text{Beta}(2, 8)$.
The label estimate $\clab{\hat{\bm{z}}}_i$ is not a random variable of the variational
model and carries no prior. For detailed prior specifications, see App.~\ref{app:sec:priors}.

\paragraph{Trait Estimation.}
We perform \gls{SVI} using \textit{Pyro}
\citep{pyro} and maximise the \gls{ELBO} conditional on the current
$\clab{\hat{\bm{z}}}$.
For details on the objective, the label step and the variational
distributions, see App.~\ref{app:sec:variational_inference}.
We base our implementation on \texttt{py-irt} \citep{lalor2023pyirt},
which was developed for fitting \gls{IRT} models on artificial crowdsourced 
data \citep{lalor2019emnlp} and improving \gls{NLP} leaderboards \citep{rodriguez2021evaluation}. 
The \gls{IRT} models are not identifiable (multiple parameterisations can
result in the same model) in an unsupervised setting, e.g.\ due to a
label-flip symmetry in the 1PL and 2PL models.
However, this can be mitigated with proper initialisation (see App.~\ref{app:sec:identifiability}).

\subsection{Multi-class \gls{IRT}}
The \gls{IRT} models described earlier act as binary classifiers, predicting the probability of annotator $j$ correctly labelling an item $i$.
An extension of this to the multi-class setting needs to define how these predictions change per class.

For each item $i$, we assume a prior distribution over the possible classes $P(\clab{Z}_i = k)$ for $k = 1, \ldots, K$,
with $K$ as the number of classes.
We assume a conditional distribution $P(\clab{L}_{ij} = k' \mid \clab{Z}_i = k)$ specifying the probability that annotator $j$ assigns class $k'$ to item $i$ given the item's real class $k$.
The diagonal elements $P(\clab{L}_{ij} = k \mid \clab{Z}_i = k)$ hold the probability of a correctly predicted label by annotator $j$ for item $i$.
Without subscript, $P$ denotes this true distribution, whereas
$P_{\bm{\psi}}(\clab{L}_{ij} = k' \mid \clab{Z}_i = k)$ denotes the
distribution defined by a model with latent traits $\bm{\psi}$, with diagonal
$p_{ij}(k) \coloneqq P_{\bm{\psi}}(\clab{L}_{ij} = k \mid \clab{Z}_i = k)$.
During trait estimation, each observation is scored with the expected
log-likelihood
$\sum_k \clab{\hat{z}}_{ik} \log P_{\bm{\psi}}(\clab{L}_{ij} \mid \clab{Z}_i = k)$,
where $\clab{\hat{z}}_{ik}$ is the soft label estimate $\clab{\hat{\bm{z}}}_i$
for class $k$ (App.~\ref{app:sec:likelihood}).

\paragraph{Uniform Extension.}
The uniform extension to the multi-class setting is to estimate a class-independent $p_{ij}(k) = p_{ij}$ for all $k$ and distribute the remaining
probability mass uniformly across the other classes
(Eq.~\ref{eq:multiclass_likelihood}). We refer to this variant as uniform \gls{IRT} in the experiments.

\begin{equation}
	P_{\bm{\psi}}(\clab{L}_{ij} = k' \mid \clab{Z}_i = k) =
	\begin{cases}
	p_{ij} & \text{if } k = k', \\
	\dfrac{1 - p_{ij}}{K - 1} & \text{otherwise}.
	\end{cases}
	\label{eq:multiclass_likelihood}
\end{equation}

\paragraph{Dawid-Skene Style Extension.}
With the uniform extension, models cannot fit class-specific confusion. However, clearly some classes are harder to distinguish than
others depending on the task. In ImageNet \citep{imagenet}, for example, a tiger shark is more easily confused with a great white than, say, with a hummingbird.

We assume a Dawid-Skene-like model of per annotator per class confusions
\citep{dawid_skene} independent of items.
We call this variant Confusion-Matrix \gls{IRT} (CM-\gls{IRT}).
A latent error matrix $\cann{\bm{E}}_j$ of annotator $j$ distributes the
remaining probability mass over the wrong classes, with
$\cann{E}_j(k, k) = 0$ and $\sum_{k'=1}^{K} \cann{E}_j(k, k') = 1$.
The diagonal follows the \gls{IRT} model, where a per-class
residual $\cann{\delta}_{jk}$ is added to the ability
$\cann{\theta}_j$ of annotator $j$
for a candidate true class $k$ (Eq.~\ref{eq:irt_cm_p}), so that
$p_{ij}(k) = p_{ij}$ if $\cann{\delta}_{jk} = 0$. The assumption behind
$\cann{\delta}_{jk}$ is that annotators recognise some classes better
than others.

\begin{equation}
	p_{ij}(k) = \citm{c}_i + (1 - \citm{c}_i)\,\sigma\left(\citm{\alpha}_i \left(\cann{\theta}_j + \cann{\delta}_{jk} - \citm{\beta}_i\right)\right)
	\label{eq:irt_cm_p}
\end{equation}

This gives Eq.~\ref{eq:cm_likelihood} with latent
traits $\bm{\psi} = (\cann{\bm{\theta}}, \cann{\bm{\delta}},
\citm{\bm{\alpha}}, \citm{\bm{\beta}}, \citm{\bm{c}}, \cann{\bm{E}})$.

\begin{equation}
\begin{aligned}
&P_{\bm{\psi}}(\clab{L}_{ij} = k' \mid \clab{Z}_i = k) \\
&\quad =
\begin{cases}
p_{ij}(k) & \text{if } k = k', \\
\bigl(1 - p_{ij}(k)\bigr)\,\cann{E}_j(k, k') & \text{otherwise}.
\end{cases}
\end{aligned}
\label{eq:cm_likelihood}
\end{equation}

For $\citm{c}_i = 0$ and residuals that sum to 0, the ability relates to the
diagonal as in Eq.~\ref{eq:diag_decomposition}.
On an item with $\citm{\alpha}_i = 1$ and $\citm{\beta}_i = 0$,
$\cann{\theta}_j$ can be interpreted as the average log-odds of being
correct (average of the diagonal).

\begin{equation}
\cann{\theta}_j = \citm{\beta}_i + \frac{1}{\citm{\alpha}_i} \cdot
\frac{1}{K} \sum_{k=1}^{K} \sigma^{-1}\bigl(p_{ij}(k)\bigr)
\label{eq:diag_decomposition}
\end{equation}

In practice, we do not constrain $\cann{\delta}_{jk}$ and use the mean of
$\cann{\theta}_j + \cann{\delta}_{jk}$ over classes as ability.

\subsection{Learning from Crowds with \gls{IRT}}
We hypothesise that the latent traits of items and annotators can be used to inform
classifiers by leveraging them using curriculum learning or forward correction.

\paragraph{Forward Correction.}
In order to introduce information about the noisy annotations into the training process, we
use the forward-correction method proposed by \citet{Patrini_2017_CVPR} with noise models based on \gls{IRT}.
Specifically, we use CM-\gls{IRT} (Eq.~\ref{eq:cm_likelihood}) as the noise model
and define $\cnoise{T}_{ijkk'} \coloneqq P_{\bm{\psi}}(\clab{L}_{ij} = k' \mid \clab{Z}_i = k)$.

\begin{figure*}[t]
  \centering
  \resizebox{\linewidth}{!}{% IRT forward-correction training: forward pass only.
% Requires: \usetikzlibrary{calc,arrows.meta}

\begin{tikzpicture}[
  font=\scriptsize,
  x=1.26437cm, % horizontal spread chosen so the figure fills \linewidth at the former 0.8\linewidth scale
  labbox/.style={
    draw=labelcolor!55, fill=labelcolor!7, rounded corners=2pt,
    minimum width=1.5cm, minimum height=0.26cm, align=center},
  featbox/.style={
    draw=clfcolor!55, fill=clfcolor!7, rounded corners=2pt,
    minimum width=1.5cm, minimum height=0.26cm, align=center},
  irtbox/.style={
    draw=noisecolor!55, fill=noisecolor!7, rounded corners=2pt,
    minimum width=1.6cm, minimum height=0.26cm, align=center},
  clfbox/.style={
    draw=clfcolor!55, fill=clfcolor!7, rounded corners=2pt,
    minimum width=1.6cm, minimum height=0.26cm, align=center},
  fwdbox/.style={
    draw=gray!50, fill=gray!10, rounded corners=2pt,
    minimum width=2.2cm, minimum height=0.26cm, align=center},
  lssbox/.style={
    draw=gray!50, fill=gray!10, rounded corners=2pt,
    minimum width=1.2cm, minimum height=0.26cm, align=center},
  arr/.style={->, >=Latex, thin},
  al/.style={font=\tiny, midway},
  alb/.style={font=\tiny, midway, fill=white, inner sep=0.5pt},
]

%--- Nodes ---
\node[labbox, anchor=west] (annot) at (0,  0.4) {Annotations $\clab{\bm{L}}_i$};
\node[featbox, anchor=west] (items) at (0, -0.4) {Features $\cclf{\bm{x}}_i$};
\node[irtbox]  (irtm)  at (3.6,  0.4) {IRT Model};
\node[clfbox]  (clfm)  at (3.6, -0.4) {Classifier $\cclf{f}_\phi$};
\node[fwdbox]  (fwdm)  at (8.9,  0)  {Correction\\[-1pt]{\tiny$\sum_k \cclf{p}_\phi(k \mid \cclf{\bm{x}}_i)\,\cnoise{T}_{ijkk'}$}};
\node[lssbox]  (lossm) at (12.6,  0) {Loss $\mathcal{L}_i$};

%--- Top route: observed labels enter loss ---
\coordinate (Ltop) at ($(annot.north) + (0, 0.38)$);
\coordinate (Rtop) at (Ltop -| lossm.north);
\draw[arr, gray!55] (annot.north) -- (Ltop)
  -- node[alb, text=black] {labels $\clab{\bm{L}}_i$} (Rtop)
  -- (lossm.north);

%--- Input arrows ---
\draw[arr] (annot) -- (irtm);
\draw[arr] (items) -- (clfm);

%--- IRT and classifier feed forward correction (Y-merge) ---
\coordinate (merge) at (7.2, 0);
\draw[-]    (irtm.east) -|
  node[alb, pos=0.08, anchor=west] {noise tensor $\cnoise{\bm{T}}_i$}
  (merge);
\draw[-]    (clfm.east) -|
  node[alb, pos=0.08, anchor=west] {predictions $\cclf{p}_\phi(\cdot \mid \cclf{\bm{x}}_i)$}
  (merge);
\draw[arr]  (merge) -- (fwdm.west);

%--- Forward correction to loss ---
\draw[arr] (fwdm) --
  node[alb] {corrected $\bm{o}_i$}
  (lossm);

\end{tikzpicture}}
  \caption{Diagram depicting the joint training approach visually. Annotations are used to fit an initial \gls{IRT} model, which is used as a noise model to correct classifier outputs, thus making 
  them comparable directly to the raw annotator labels. Both the classifier and the \gls{IRT} model are updated from the loss.}
  \label{fig:joint_irt_training}
\end{figure*}
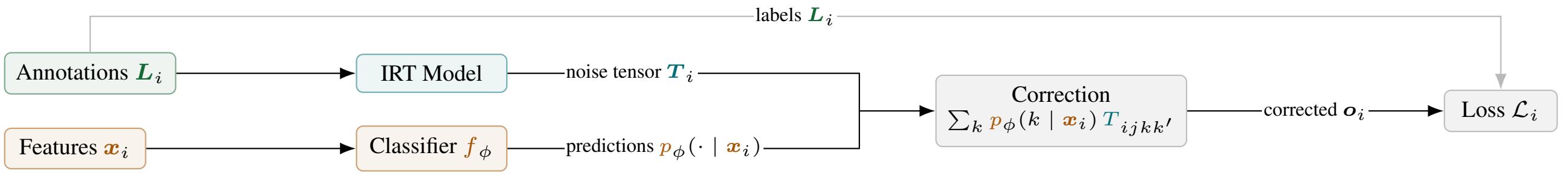

The model's predicted class distribution $\cclf{p}_\phi(k \mid \cclf{\bm{x}}_i) = \text{softmax}(\cclf{f}_\phi(\cclf{\bm{x}}_i))_k$
is then corrected (Eq.~\ref{eq:forward_cm_loss}).
That is, we sum over true-class ($k$) indices to get an estimate for observed annotations.
We can then calculate the cross-entropy loss directly on the individual annotator labels (Eq.~\ref{eq:forward_cm_loss}), since we 
applied correction for label noise beforehand.

\begin{equation}
	\begin{gathered}
	o_{ijk'} = \sum_{k} \cclf{p}_\phi(k \mid \cclf{\bm{x}}_i)\,\cnoise{T}_{ijkk'}, \\
	\mathcal{L} = -\frac{1}{\sum_{i,j} m_{ij}}\sum_{i,j} m_{ij} \log o_{ij \clab{L}_{ij}}
	\end{gathered}
	\label{eq:forward_cm_loss}
\end{equation}

where $m_{ij} \in \{0, 1\}$ masks annotator-item pairs without an observed vote, and $\clab{L}_{ij}$ indexes
the class annotator $j$ reported for item $i$.
Rather than freezing $\cnoise{\bm{T}}$, we let the classifier's belief about an item
inform the annotation noise model. Following \citet{Rodrigues_Pereira_2018} we
perform joint training of the annotation noise model and the classifier
(Fig.~\ref{fig:joint_irt_training}). \gls{IRT} parameters are warm-started
using an \gls{SVI}-fitted CM-\gls{IRT} model.
Then, the annotator traits $\cann{\theta}_j$, $\cann{\delta}_{jk}$ and the
classifier parameters are updated jointly at each training step, with a
single gradient step on the shared loss in Eq.~\ref{eq:forward_cm_loss}.
The latent error matrix is not trained but set to uniform errors,
since using the fitted error distribution led to unstable training.
The item traits $\citm{\beta}_i$, $\citm{\alpha}_i$ and $\citm{c}_i$ are kept at their
warm-start values and shifted by residuals predicted from the classifier's
embedding of $\cclf{\bm{x}}_i$, such that they can only vary through the features of the item.

\begin{table*}[t]
\caption{Overview of crowd-annotations on the crowd-annotated training
split of all datasets.}
\label{tab:dataset-overview}
\resizebox{\linewidth}{!}{
\begin{tabular}{lrrrrrr}
\toprule
\textbf{Dataset} & \textbf{Annotators} & \textbf{Samples} & \textbf{Labels per Sample} & \textbf{Labels per Annotator} & \textbf{MV Acc. (\%)} & \textbf{Mean Ann. Acc. (\%)}\\
\midrule
Dopanim & 20 & 10,484 & 4.97 $\pm$ 0.19 & 2602.80 $\pm$ 1254.96 & 80.81 & 65.57 $\pm$ 14.56 \\
MusicGenres & 44 & 700 & 4.21 $\pm$ 2.00 & 66.93 $\pm$ 105.62 & 69.71 & 73.28 $\pm$ 24.43 \\
RTE & 164 & 560 & 10.00 $\pm$ 0.00 & 34.15 $\pm$ 78.74 & 89.29 & 83.56 $\pm$ 12.43 \\
SentimentPolarity & 203 & 4,999 & 5.55 $\pm$ 0.69 & 136.68 $\pm$ 346.22 & 88.50 & 77.12 $\pm$ 17.15 \\
CIFAR10H & 2571 & 10,000 & 51.42 $\pm$ 1.48 & 200.00 $\pm$ 0.00 & 99.22 & 94.87 $\pm$ 5.73 \\
CIFAR10N & 747 & 50,000 & 3.00 $\pm$ 0.00 & 200.80 $\pm$ 329.30 & 91.16 & 82.06 $\pm$ 6.16 \\
\bottomrule
\end{tabular}}
\end{table*}

\paragraph{Curriculum Learning.}
We apply the method introduced by \citet{lalor-yu-2020-dynamic} in Sec.~\ref{sec:dds} for curriculum learning and change the sampling process for training batches to always sample items at or below the
current classifier's ability. Before training, an \gls{IRT} model is fitted to the annotations using SVI. During training, the classifier ability ($\cann{\theta}_{\phi}$)
is estimated per epoch using a simple \gls{MAP} estimate by keeping item traits fixed and solving for $\cann{\theta}_{\phi}$
using 5 steps of a Newton-Raphson solver, where the normal prior is set according to the mean and standard deviation of abilities in the initial \gls{IRT} fit.

\section{Experiments}

\begin{figure*}[t]
	\centering
	\subfloat[\label{fig:trait_corrs_fig}]{%
		\includegraphics[width=0.90\linewidth]{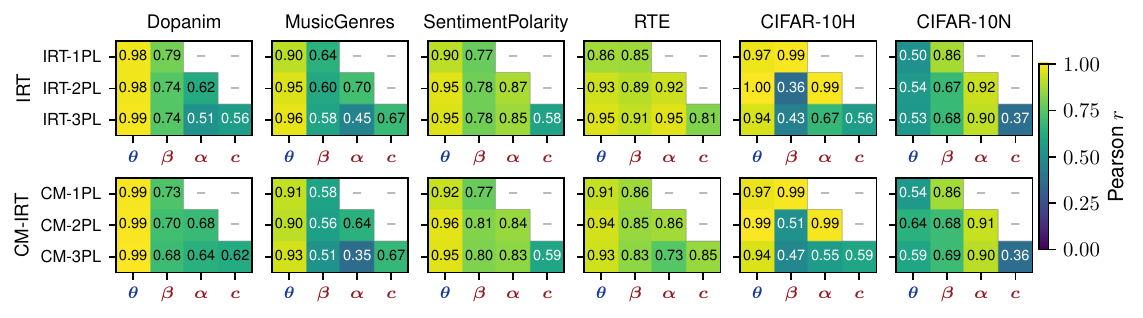}}\\
	\subfloat[\label{fig:benchmark_barplots}]{%
		\includegraphics[width=0.95\linewidth]{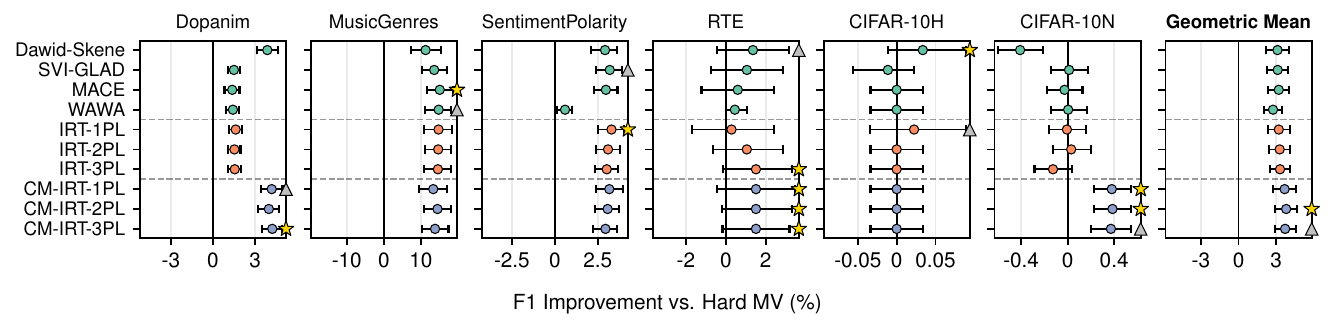}}
	\caption{(a)~Pearson correlation between latent traits recovered by unsupervised
	\gls{IRT} models and their supervised counterparts fitted on
	ground truth labels, across all datasets.
	(b)~Label-inference benchmark F1 improvement relative to
	hard~\gls{MV} across all datasets.
	A gold star marks the best model per dataset and
	a silver triangle marks the second-best.
	Error bars indicate the 95\% CI calculated by 1k paired bootstrap
	resamples of the difference against hard~\gls{MV}.
	Brier Score results are in the appendix
	(Fig.~\ref{fig:benchmark_barplots_brier}) and the
	complete results in Tab.~\ref{tab:benchmark_results}.}
	\label{fig:trait_and_benchmark}
\end{figure*}

We initially confirm that our method can successfully recover the latent traits (RQ1) and true
labels (RQ2) by evaluating on a set of benchmarks.
We then investigate the utility of the inferred traits for learning from
crowds by using them to inform the training of a downstream model (RQ3).

\subsection{Crowd-Labelled Datasets}
We evaluate our methods on various datasets from the image, text and audio
modalities in binary or multi-class settings.
All datasets provide expert-confirmed ground truth labels independent of the
crowd labels (more in App.~\ref{app:sec:datasets}).

\begin{itemize}
    \item \textbf{Dopanim} \citep{dopanim} contains around 16k animal images
    annotated by 20 laypeople, featuring doppelganger animals of visually
    similar species as a challenging benchmark.

    \item \textbf{MusicGenres} \citep{RODRIGUES20131428} contains 1k audio
    samples from 10 music genres, annotated by 44 workers, using pre-extracted
    timbral and spectral features \citep{RODRIGUES20131428}.

    \item \textbf{SentimentPolarity} \citep{pang-lee-2005-seeing} contains
    sentiment annotations from \gls{AMT} workers for movie reviews from
    RottenTomatoes \citep{RODRIGUES20131428}.

    \item \textbf{\gls{RTE}} \citep{rte} is a natural language inference
    dataset where workers label whether premise-hypothesis pairs show entailment.

    \item \textbf{CIFAR10H} \citep{cifar10h} and \textbf{CIFAR10N}
    \citep{cifar10n} provide crowd annotations for CIFAR10 \citep{cifar10},
    with ${\sim}51$ and ${\sim}3$ labels per image, respectively.
\end{itemize}

For all datasets except RTE, only the training split was annotated by the
crowd, while validation and test splits only contain ground truth labels.
We hold out 10\% of the crowd-annotated training data for validation and run
all label inference and trait recovery experiments on the remaining 90\%.
For \gls{RTE}, where all items are crowd-annotated, we run both on the
training and test splits.
For an overview of the properties of annotations, please refer to Tab.~\ref{tab:dataset-overview}.
We split \gls{RTE} into 70\% training, 10\% validation and 20\% test data,
and reuse the same training and test splits as \citet{Herde2024} for all
other datasets.
For an overview of the splits, please refer to Tab.~\ref{tab:dataset_splits} in the appendix.

\subsection{Unsupervised Trait Inference} \paragraph{Evaluation.} In order to test how well unsupervised \gls{IRT} models recover the latent
traits of items, we compare them against supervised \gls{IRT} variants fitted
using the ground truth labels for $\clab{Z}_i$ instead of $\clab{\hat{\bm{z}}}_i$.
We test both uniform \gls{IRT} models and confusion-modelling CM-\gls{IRT} variants.

\paragraph{Results.}
Pearson correlations between supervised and unsupervised traits on all datasets are
shown in Fig.~\ref{fig:trait_corrs_fig}.
Annotator ability is recovered reliably across all three parameterisations, except on CIFAR10N. 
This indicates that the unsupervised models identify the
relative competence of annotators almost as well as their supervised counterparts.
Item difficulty is recovered at a lower precision and less consistently,
especially by the 2PL and 3PL models on CIFAR10H.
Recovery of the item-specific traits introduced by richer models is
harder: discrimination correlations drop on average in the 2PL and 3PL models.
The guessing floor seems to be hardest to recover overall.
Recovery seems to be similar for CM-\gls{IRT} models when compared to
uniform \gls{IRT} models.

\begin{figure*}[!t]
	\centering
	\includegraphics[width=0.95\linewidth]{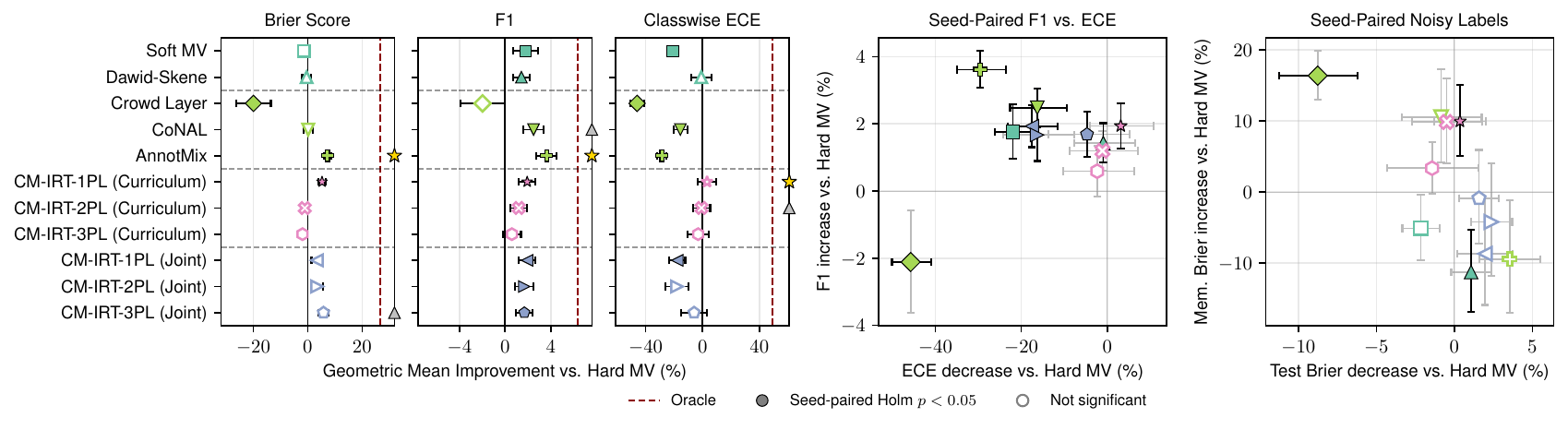}
	\caption{Aggregated results of the crowd-learning experiments across all datasets.
	The left panels show the improvement of metrics over the hard~\gls{MV} baseline.
	Error bars indicate 95\% CIs based on pooled per-dataset standard
	errors ($t$-intervals, App.~\ref{app:sec:perm_test}).
	The rightmost panel shows the memorisation experiment
	with noisy labels.
	Filled markers indicate a significant difference to hard~\gls{MV} in a
	seed-paired permutation test (Holm-adjusted $p < 0.05$,
	App.~\ref{app:sec:perm_test}).
	The per-dataset results can be found in the appendix
	(Fig.~\ref{fig:classifier_barplots}, Tab.~\ref{tab:classifier_results}).}
	\label{fig:classifier_aggregate}
\end{figure*}

\subsection{Label Inference}

\paragraph{Evaluation.} We evaluate how well \gls{IRT} models can recover ground
truth labels from crowdsourced annotations in our benchmark datasets.
We run the label inference methods on annotator labels only and compare the inferred labels 
to ground truth labels using F1 and the Brier Score. The Brier Score is a scoring rule
that measures the quality of probabilistic predictions by the mean squared error between the
predicted probabilities and the actual outcomes.
This is informative, since it captures not only the accuracy of the predictions, but also their calibration.

\paragraph{Baselines.}
We use \texttt{crowd-kit} \citep{CrowdKit} to compare against the
established label inference algorithms described in
App.~\ref{app:sec:label_inference_baselines}.
Specifically, we compare against Majority Vote (hard and soft), WAWA,
Dawid-Skene \citep{dawid_skene}, MACE \citep{mace}
and SVI-GLAD \citep{glad}.

\paragraph{Results.} 
We observe that \gls{IRT} models, specifically CM-\gls{IRT}, are the 
most stable models over all datasets (see geometric mean in Fig.~\ref{fig:benchmark_barplots}).
On Dopanim, we can distinctly see the advantage of modelling confusions by the gains of CM-\gls{IRT} models over uniform \gls{IRT} 
and by the strong performance of the Dawid-Skene model. Recall that the dataset features highly confusable classes.  
Further analysis of model robustness to adversarial annotators and
performance as a function of the number of raters per item is provided
in App.~\ref{app:sec:adv_raters}.

\subsection{Learning from Crowds}

\paragraph{Setup.}
We use \texttt{maml} \citep{herde_maml}
as a basis for our experiments on learning from crowds.
Following \citet{Herde2025crowdhpoRH}, we use 10\% of the training data for
validation instead of ground-truth validation sets used by \citet{Herde2024}, making
our scores not directly comparable to those reported there.
We report the mean and standard error of the mean across 10 seeds for all
classifiers and datasets. Controlled synthetic experiments confirming that \gls{IRT}-informed classifiers
outperform crowd baselines under fulfilled model assumptions are provided in
App.~\ref{app:sec:synthetic_testbed}.

\paragraph{Backbone Models.}
We use an \gls{MLP} with residual connections for tabular datasets
(MusicGenres, SentimentPolarity), DINOv2-base \citep{dinov2} for Dopanim,
ResNet-18 \citep{resnet18} for CIFAR10H/N, and DistilBERT
\citep{distilbert} fine-tuned on \gls{MNLI} \citep{mnli} for \gls{RTE}.
Each backbone is followed by a method-specific output layer. Details and
hyperparameters are in the appendix (Tab.~\ref{tab:output_configs}).
This is largely the setup used by \citet{Herde2024} apart from tabular datasets.

\paragraph{Baselines.}
We compare against the following methods for learning from crowds,
all implemented via \texttt{maml} \citep{Herde2024}: aggregate classifiers (majority-voted hard labels,
soft majority-vote, Dawid-Skene soft labels), Crowd Layer
\citep{Rodrigues_Pereira_2018}, CoNAL \citep{Chu_Ma_Wang_2021},
and AnnotMix \citep{Herde2024}. The details are in
App.~\ref{app:sec:lfc_baselines}.

\paragraph{Performance across Datasets.} We aggregate results over all datasets by the geometric mean of the percentage gain 
over the hard~\gls{MV} baseline, such that the different difficulties of the datasets do not bias the aggregate. 
Fig.~\ref{fig:classifier_aggregate} shows a forest plot of the different mean performances for all methods. 
We find that AnnotMix reaches the best performance in terms of F1 in general, 
but appears miscalibrated in terms of class-wise \gls{ECE} \citep{cece}.
Soft \gls{MV} performs decently in terms of F1, but is miscalibrated as well. 
Joint CM-\gls{IRT} methods outperform non-crowd-learning baselines on Brier Score, and the 1PL variant also on F1, but seem worse in terms of calibration.
Curriculum learning approaches are better calibrated, and the 1PL variant still outperforms non-crowd-learning baselines on F1 and Brier Score.

\paragraph{Significance Testing.}
We test each method against hard~\gls{MV} with a seed-paired permutation test
with Holm correction \citep{holm} (see App.~\ref{app:sec:perm_test}) due to multiple tests.
Significant differences are indicated by filled markers in Fig.~\ref{fig:classifier_aggregate}.
We notice that no method is significantly better than hard~\gls{MV} in both F1 and ECE.
The 1PL curriculum learning variant is the only approach with improvement in both F1 and ECE, but it is non-significant in terms of ECE.

\paragraph{Memorisation Experiments.}
We hypothesise that models trained on crowd-data generalise better
and exhibit less memorisation of noisy labels than models trained on
majority-voted labels.
To test this, we conduct an experiment on MusicGenres and RTE, where we
randomise the labels of 10\% of the training items by flipping (binary case)
or rotating (multi-class case, random offset) and then evaluate the Brier Score
on the noisy training items and the clean test set. We first select a new label
for each item by uniformly sampling from the set of possible labels, excluding
the original label.
We then shift all annotator labels on the item by the same offset, such that
correct annotator labels become the new label and incorrect labels are rotated
accordingly.
If a model is memorising the noisy labels, we expect a lower Brier Score on
the noisy training items, measured against the new labels. We compare memorisation vs.\ test-set Brier Scores,
hypothesising that crowd-learning methods will reduce memorisation of noisy examples.

\paragraph{Results.}
The rightmost panel of Fig.~\ref{fig:classifier_aggregate} shows memorisation
vs.\ test performance for all models under the Brier Score.
Similar to the F1 and ECE comparison, we find that the only method positive on both $x$ and $y$ axes 
is the 1PL curriculum learning method. This means it is the only approach that both decreases
memorisation of noisy labels and improves the test-set Brier Score. However, the improvement on the test set is not significant.

\section{Conclusion}
We show how \gls{IRT} models can be used in an unsupervised setting for
inferring interpretable item traits (RQ1) and true labels (RQ2) from
crowdsourced data. \gls{IRT} models seem to be the most stable label-inference
models across the evaluated datasets and provide the best estimates in terms
of F1 for 4 out of 6 benchmarks.
However, the margins to the next-best baseline are small.

Furthermore, we find that using item traits or crowd-labels to inform training can improve performance in terms of F1 and Brier Score when compared to a majority-vote (RQ3).
However, \gls{IRT}-informed methods do not reach state-of-the-art performance
in crowd learning, but the curriculum variants seem to preserve calibration
better than other crowd-learning methods.
Future research should test these findings under hyperparameter-tuning.

\section{Limitations}
While we do our best to make a fair comparison to existing crowd-learning methods, we do not perform hyperparameter tuning due to computational limitations,
which has been shown to change results substantially in some cases \citep{Herde2025crowdhpoRH}. More work is needed to validate the observed trade-off between calibration and F1 in crowd-learning methods under hyperparameter optimisation.

\section*{Impact Statement}
We comply with data licenses by strictly using data for research and not
providing them directly with our code, which will be made available upon
peer-reviewed publication. Rather, they must be obtained from
the source. Furthermore, item traits obtained by \gls{IRT} models could be
biased if they correlate with the demographics of annotators. Annotator ability
could be biased for annotators holding valid, but non-majority views.
Care should be taken when leveraging item and annotator traits in settings
where such biases could occur.
This paper presents work whose goal is to advance the field of machine
learning. Beyond the above, there are many potential societal consequences of
our work, none of which we feel must be specifically highlighted here.

\section*{AI Disclosure Statement}
We used Claude Code (models: Claude Sonnet 4.6, Claude Sonnet 5, Claude Opus 5,
and Claude Opus 5.5) for executing experiments (code generation) and writing
assistance (editing) of this work under close supervision.
We verified code and outputs by continuous review.
Initial drafts for the article were written by the authors and subsequently
edited and reviewed with Claude Code.

\bibliography{custom}
\bibliographystyle{icml2026}
\clearpage

\appendix
\section{Reproducibility Statement}
The complete source code (including random seeds, experiment configurations
and hyperparameterisations) for reproducing the experimental results will be
made available on GitHub upon peer-reviewed publication.
We will also provide experimental data (e.g., calculated statistics, per-sample and seed predictions of models) together with the code, which should aid in 
repeating hypothesis tests without needing to rerun all experiments. Public datasets released
by other parties are not provided directly and must be obtained from the source.

\section{Unsupervised \gls{IRT}}

\subsection{Why the EM-Style Hybrid Approach?}
Using SVI, we could jointly estimate $\clab{\hat{\bm{z}}}_i$ together with the \gls{IRT} traits. 
However, in practice we observed that the choice of prior for $\clab{\hat{\bm{z}}}_i$ is problematic and
has a large influence on the calibration of the inferred labels compared to ground truth.
Which prior would work was also dependent on the dataset.
To avoid this problem, we simply alternate between solving for $\clab{\hat{\bm{z}}}_i$ analytically and 
optimising \gls{IRT} traits with \gls{SVI}. 
This way, we do not have to settle on a prior for $\clab{\hat{\bm{z}}}_i$.

\subsection{Unsupervised \gls{IRT} Identifiability}
\label{app:sec:identifiability}
\gls{IRT} models are not identifiable in an unsupervised setting.
This means multiple parameter configurations can
explain the data equally well, making unique parameter recovery
impossible.
For instance, the 2PL model exhibits a symmetry where flipping
the label ($1 - \clab{Z}_i$), negating abilities ($-\cann{\theta}_j$), and
negating difficulties ($-\citm{\beta}_i$) results in an equivalent model
(Eq.~\ref{eq:symmetry}), where $\bm{\psi}'$ equals $\bm{\psi}$ with
$\cann{\theta}_j$ and $\citm{\beta}_i$ negated.
However, with sensible initialisation by majority-vote and validation
after fitting, we can virtually eliminate the risk of finding a
flipped solution.
A flip would likely occur when more than half of the labels are wrong,
which would render the crowdsourced data uninformative in the
first place.

\begin{align}
	P_{\bm{\psi}'}(\clab{L}_{ij} = l \mid 1 - \clab{Z}_i)	\label{eq:symmetry}\\
	= P_{\bm{\psi}}(\clab{L}_{ij} = l \mid \clab{Z}_i)\nonumber
\end{align}

For CM-\gls{IRT}, only the sum $\cann{\theta}_j + \cann{\delta}_{jk}$
is used in the likelihood, so $\cann{\theta}_j$ and $\cann{\delta}_{jk}$ are not
identifiable individually without the constraint $\sum_k \cann{\delta}_{jk} = 0$.

\subsection{Prior Specifications}
\label{app:sec:priors}

For ability ($\cann{\theta}_j$), difficulty ($\citm{\beta}_i$), and discrimination
($\citm{\alpha}_i$), we use hierarchical priors.
Each group has an independent mean hyperprior
$\mu \sim \mathcal{N}(0,1)$ and scale hyperprior
$\tau \sim \text{HalfNormal}(1)$, with parameters drawn
conditionally from these group-level distributions.
The guessing parameter uses a fixed distribution.

\begin{align}
	\mu_{\cann{\theta}}, \mu_{\citm{\beta}}, \mu_{\citm{\alpha}} &\sim \mathcal{N}(0,1)\label{eq:priors} \\
	\tau_{\cann{\theta}}, \tau_{\citm{\beta}}, \tau_{\citm{\alpha}} &\sim \text{HalfNormal}(1)\nonumber \\
	\cann{\theta}_j &\sim \mathcal{N}(\mu_{\cann{\theta}}, \tau_{\cann{\theta}})\nonumber \\
	\citm{\beta}_i &\sim \mathcal{N}(\mu_{\citm{\beta}}, \tau_{\citm{\beta}})\nonumber \\
	\citm{\alpha}_i &\sim \text{LogNormal}(\mu_{\citm{\alpha}}, \tau_{\citm{\alpha}})\nonumber \\
	\citm{c}_i &\sim \text{Beta}(2,8)\nonumber
\end{align}

For CM-\gls{IRT}, the per-class ability residuals and the error matrices
receive the priors
$\tau_{\cann{\delta},k} \sim \text{HalfNormal}(1)$,
$\cann{\delta}_{jk} \sim \mathcal{N}(0, \tau_{\cann{\delta},k})$ and
$\cann{\bm{E}}_j(k, \cdot) \sim \text{Dirichlet}(\bm{1}_{K-1})$
over the $K-1$ off-diagonal entries of each row.
The label estimate $\clab{\hat{\bm{z}}}_i$ has no prior in the variational model,
since it is not a latent variable of $q$
(App.~\ref{app:sec:variational_inference}).
In the multi-class setting ($K > 2$), observations are modelled with a
Categorical distribution instead of Bernoulli.

\subsection{SVI Likelihood}
\label{app:sec:likelihood}
The item response models define the probability of a correct response as
$p_{ij} = P_{\bm{\psi}}(\clab{L}_{ij} = \clab{Z}_i)$.
Given a hard true label $\clab{Z}_i \in \{0, 1\}$, the likelihood of an
observed label $\clab{L}_{ij} \in \{0, 1\}$ is expressed as in
Eq.~\ref{eq:likelihood}, with $\mathbbm{1}[.]$ as the indicator
function that evaluates to 1 if the condition holds, and otherwise
to 0.

\begin{align}
	P_{\bm{\psi}}(\clab{L}_{ij} \mid \clab{Z}_i) =\label{eq:likelihood} \\
	\mathbbm{1}[\clab{L}_{ij} = \clab{Z}_i] p_{ij} + \mathbbm{1}[\clab{L}_{ij}		\neq \clab{Z}_i] (1 - p_{ij}) \nonumber
\end{align}

In the trait step, $\clab{Z}_i$ is replaced by the soft estimate
$\clab{\hat{z}}_i = P_{\bar{\bm{\psi}}}(\clab{Z}_i = 1 \mid \clab{\bm{L}}_i) \in [0, 1]$ from the
label step (Eq.~\ref{eq:label_step_binary}).
Since $\clab{\hat{z}}_i$ is a posterior that already includes $\clab{\bm{L}}_i$,
it is not mixed into the likelihood as a prior over $\clab{Z}_i$.
Instead, as in the M-step of EM, each observation is scored with the
expected log-likelihood under $\clab{\hat{z}}_i$
(Eq.~\ref{eq:expected_loglik}), where both terms are given by
Eq.~\ref{eq:likelihood}.

\begin{equation}
	\begin{split}
		\ell_{ij} = {} & \clab{\hat{z}}_i \log P_{\bm{\psi}}(\clab{L}_{ij} \mid \clab{Z}_i = 1) \\
		& + (1 - \clab{\hat{z}}_i) \log P_{\bm{\psi}}(\clab{L}_{ij} \mid \clab{Z}_i = 0)
	\end{split}
	\label{eq:expected_loglik}
\end{equation}

In the multi-class setting ($K > 2$), the observation model changes
to a categorical distribution.
Here $\clab{\hat{\bm{z}}}_i \in \Delta^{K-1}$ (the probability simplex over
$K$ classes) is the soft label from the
label step (Eq.~\ref{eq:label_step_multiclass}), with component
$\clab{\hat{z}}_{ik} = P_{\bar{\bm{\psi}}}(\clab{Z}_i = k \mid \clab{\bm{L}}_i)$.
The probability $P_{\bm{\psi}}(\clab{L}_{ij} \mid \clab{Z}_i = k)$ of observing $\clab{L}_{ij}$
given true class $k$ is given by Eq.~\ref{eq:multiclass_likelihood}, where incorrect
probability mass is spread equally over the $K-1$ wrong classes
(or by Eq.~\ref{eq:cm_likelihood} for CM-\gls{IRT},
which additionally depends on $\cann{\delta}_{jk}$ and
$\cann{\bm{E}}_j(k, \cdot)$).
Each observation is then scored with the expected log-likelihood
over all $K$ classes (Eq.~\ref{eq:expected_loglik_multiclass}),
which reduces to Eq.~\ref{eq:expected_loglik} for $K = 2$.

\begin{equation}
	\ell_{ij} = \sum_{k=1}^{K} \clab{\hat{z}}_{ik}
	\log P_{\bm{\psi}}(\clab{L}_{ij} \mid \clab{Z}_i = k).
	\label{eq:expected_loglik_multiclass}
\end{equation}

The
expected log-likelihood over the set of all observed item-annotator
pairs $\mathcal{O}$ is shown in Eq.~\ref{eq:likelihood_full}.
It takes the place of the log-likelihood in the \gls{ELBO}
(Eq.~\ref{eq:elbo}).
For a one-hot $\clab{\hat{\bm{z}}}$ it equals the log-likelihood of
Eq.~\ref{eq:likelihood}.

\begin{equation}
	\log P_{\bm{\psi}}(\clab{\bm{L}} \mid \clab{\hat{\bm{z}}}) \coloneqq \sum_{(i,j) \in \mathcal{O}} \ell_{ij}.
	\label{eq:likelihood_full}
\end{equation}

\subsection{Variational Distributions and Initialisation}
\label{app:sec:variational_distributions}

We use a normal posterior for $\cann{\theta}_j$ and $\citm{\beta}_i$, a log-normal
posterior for $\citm{\alpha}_i$, and a beta posterior for $\citm{c}_i$.
For CM-\gls{IRT}, we additionally use a normal posterior for $\cann{\delta}_{jk}$
and a Dirichlet posterior
$q(\cann{\bm{E}}_j(k, \cdot)) = \text{Dirichlet}(\bm{\kappa}_{jk})$
with concentrations $\bm{\kappa}_{jk}$ initialised to one.
There is no variational distribution over $\clab{\hat{\bm{z}}}_i$.
To mitigate the identifiability problem described in
App.~\ref{app:sec:identifiability}, we initialise both
$\clab{\hat{\bm{z}}}$ and the posteriors using the majority vote
of the labels.
In the binary case, the majority vote $v_i \in \{0, 1\}$ is clipped
away from the boundary to obtain the initial soft label
(Eq.~\ref{eq:init_z}). In the multi-class case, $\clab{\hat{\bm{z}}}_i$
is initialised as the one-hot encoding of the plurality class.
The first label step replaces this initialisation after one trait step.
For $\cann{\theta}_j$, we use the accuracy $a_j$ of annotator $j$ with respect
to the majority vote for initialisation (Eq.~\ref{eq:post_theta}).
This results in an initial ability ($\cann{\theta}_j$) between -1
(accuracy = 0) and +1 (accuracy = 1).
The difficulty $\citm{\beta}_i$ posterior is initialised using the sample
variance of votes for item $i$, centred by subtracting the mean
sample variance across items (Eq.~\ref{eq:post_beta}).
The guessing parameter and the discrimination remain unaffected by the
majority vote initialisation (Eq.~\ref{eq:post_guess} and
Eq.~\ref{eq:post_alpha}).

\begin{align}
	\clab{\hat{z}}_i^{(0)} & = \min\left(\max\left(v_i, 0.05\right), 0.95\right) \label{eq:init_z}                                                                 \\[0.5em]\medskip
	q(\cann{\theta}_j)  & = \mathcal{N}(2 \cdot (a_j - 0.5), 1) \label{eq:post_theta}                                                                               \\[0.5em]\medskip
	q(\citm{\beta}_i)   & = \mathcal{N}\left(\frac{\sum_{j \in \mathcal{J}_i} (\clab{L}_{ij} - \clab{\bar{L}}_i)^2}{|\mathcal{J}_i| - 1} - \overline{s^2}, 1\right) \label{eq:post_beta} \\[0.5em]\medskip
	q(\citm{c}_i)       & = \text{Beta}(1, 1) \label{eq:post_guess}                                                                                                 \\[0.5em]\medskip
	q(\citm{\alpha}_i)  & = \text{LogNormal}(0, 1) \label{eq:post_alpha}
\end{align}

\subsection{Variational Inference}
\label{app:sec:variational_inference}

We fit the model with a variational-EM hybrid that alternates a
closed-form label step with a stochastic variational trait step.
Let $\bm{\psi} = (\cann{\bm{\theta}}, \citm{\bm{\alpha}},
\citm{\bm{\beta}}, \citm{\bm{c}})$ denote the latent traits
(plus $\cann{\bm{\delta}}$ and $\cann{\bm{E}}$ for CM-\gls{IRT}).

\paragraph{Trait Step.}
With $\clab{\hat{\bm{z}}}$ held fixed, we maximise the \gls{ELBO}
(Eq.~\ref{eq:elbo}) using Monte Carlo estimates of the gradient
\citep{pmlr-v33-ranganath14}, since the expectation is analytically
intractable.
The label estimate $\clab{\hat{\bm{z}}}$ receives no gradient in this step.

\begin{align}
	\label{eq:elbo}
	\text{ELBO}(q; \clab{\hat{\bm{z}}}) = \mathbb{E}_{q(\bm{\psi})}[\log P_{\bm{\psi}}(\clab{\bm{L}} \mid \clab{\hat{\bm{z}}})] - \\
	D_{KL}(q(\bm{\psi}) \| p(\bm{\psi}))\nonumber
\end{align}

We use the mean-field approximation, where we assume the variational
distribution factorises as shown in Eq.~\ref{eq:vardist}.
For CM-\gls{IRT}, the product additionally contains
$\prod_{j,k} q(\cann{\delta}_{jk})\, q(\cann{\bm{E}}_j(k, \cdot))$.

\begin{equation}
	\label{eq:vardist}
	q(\bm{\psi}) = \prod_j q(\cann{\theta}_j) \prod_i q(\citm{\beta}_i) \prod_i q(\citm{\alpha}_i) \prod_i q(\citm{c}_i)
\end{equation}

\paragraph{Label Step.}
Before every trait step but the first, $\clab{\hat{\bm{z}}}$ is replaced by the posterior
over the true label given the votes $\clab{\bm{L}}_i$ on item $i$, with a
uniform prior over classes.
The traits are fixed at point estimates $\bar{\bm{\psi}}$
from the current $q$: the posterior means of $\cann{\theta}_j$, $\citm{\beta}_i$,
$\cann{\delta}_{jk}$, $\citm{c}_i$ and $\cann{\bm{E}}_j(k, \cdot)$, and the posterior
median $\exp(\mu^q_{\citm{\alpha}_i})$ of the log-normal $\citm{\alpha}_i$,
which does not depend on the posterior uncertainty.
Since the prior is uniform, it cancels, and the binary and multi-class
label posteriors are

\begin{align}
	\clab{\hat{z}}_i &= \sigma\left(\sum_{j : (i,j) \in \mathcal{O}}
	\log \frac{P_{\bar{\bm{\psi}}}(\clab{L}_{ij} \mid \clab{Z}_i = 1)}
	{P_{\bar{\bm{\psi}}}(\clab{L}_{ij} \mid \clab{Z}_i = 0)}\right),
	\label{eq:label_step_binary} \\
	\clab{\hat{z}}_{ik} &= \operatorname{softmax}_k\left(\sum_{j : (i,j) \in \mathcal{O}}
	\log P_{\bar{\bm{\psi}}}(\clab{L}_{ij} \mid \clab{Z}_i = k)\right),
	\label{eq:label_step_multiclass}
\end{align}

where $P_{\bar{\bm{\psi}}}(\clab{L}_{ij} \mid \clab{Z}_i = k)$ is the
observation model of Eq.~\ref{eq:likelihood} (binary),
Eq.~\ref{eq:multiclass_likelihood}
(multi-class), or Eq.~\ref{eq:cm_likelihood}
(CM-\gls{IRT}).
One label step followed by one full-batch trait step forms one epoch.

\section{Datasets}
\label{app:sec:datasets}

\begin{itemize}

    \item \textbf{Dopanim} \citep{dopanim} contains around 16k animal images
    with ground truth labels, annotated by 20 laypeople.
    The dataset contains doppelganger animals, which are visually similar
    animals of different species, making it a challenging benchmark.

    \item \textbf{MusicGenres} \citep{RODRIGUES20131428} contains 1k 30-second
    samples from 10 genres (classical, country, disco, hiphop, jazz, rock,
    blues, reggae, pop, metal), annotated by 44 workers.
    We adopt the feature set from \citep{RODRIGUES20131428}, comprising the
    means and variances of timbral descriptors, zero-crossing rate, spectral
    centroid, spectral rolloff, spectral flux, and \glspl{MFCC}.

    \item \textbf{SentimentPolarity} \citep{pang-lee-2005-seeing} contains
    more than 10k movie reviews from RottenTomatoes.
    Each review was annotated as ``Fresh'' (positive) or ``Rotten'' (negative).
    Subsequent work \citep{RODRIGUES20131428} annotated half the reviews via
    \gls{AMT} workers and the other half was kept for evaluation.

    \item \textbf{\gls{RTE}} \citep{rte} is a natural language inference
    dataset where workers decide whether a hypothesis is entailed by a premise.
    It consists of roughly 50\% positive (entailed) and 50\% negative (not
    entailed) pairs.

    \item \textbf{CIFAR10H} \citep{cifar10h} and \textbf{CIFAR10N}
    \citep{cifar10n} provide human annotations for CIFAR10 \citep{cifar10}
    gathered via \gls{AMT}.
    CIFAR10N contains 50k annotated training images with ${\sim}3$ labels per
    image. CIFAR10H contains 10k test-set images with ${\sim}51$ labels per
    image.

\end{itemize}

\section{Baselines}

\subsection{Label Inference Baselines}
\label{app:sec:label_inference_baselines}

We use \texttt{crowd-kit} \citep{CrowdKit} to compare against
the following established label inference algorithms.

\begin{itemize}
    \item \textbf{\gls{MV}:} The hard majority vote assigns
    the label with the highest frequency among annotators, assuming all
    annotators have uniform reliability.
    The soft variant produces a soft label by normalising the per-class
    annotation counts, such that the distribution over classes sums
    to~1. The hard vote is therefore its argmax.
    \item \textbf{WAWA:} A weighted majority voting approach where
    annotators' weights are iteratively adjusted based on their
    agreement with the current consensus \citep{CrowdKit}.
    This method was introduced by the \texttt{crowd-kit} authors.
    \item \textbf{Dawid-Skene:} A probabilistic model that uses the
    \gls{EM} algorithm to estimate the true labels by modelling each
    annotator's reliability through a confusion matrix
    \citep{dawid_skene}.
    \item \textbf{GLAD:} An \gls{IRT}-like logistic model that accounts
    for two latent traits: the individual annotators' ability and the
    specific difficulty of the task \citep{glad} as a multiplicative
    factor.
    We fit our own implementation of GLAD with SVI, since the
    \texttt{crowd-kit} implementation lacks multi-class support.
    \item \textbf{MACE:} An item-response model designed to identify
    spammers by assuming that annotators either behave competently
    (providing the true label) or provide noise \citep{mace}.
\end{itemize}

\subsection{Learning from Crowds Baselines}
\label{app:sec:lfc_baselines}

We compare against the following methods for learning from crowds,
all implemented via \texttt{maml} \citep{Herde2024}.

\begin{itemize}
    \item \textbf{Aggregate Classifiers:} We train classifiers on
    majority-voted hard labels as well as soft labels generated by
    soft majority-vote or a Dawid-Skene model.
    Classifiers are trained with a cross-entropy loss that natively
    supports soft labels.
    \item \textbf{Crowd Layer:} An end-to-end approach that places a
    per-annotator noise layer on top of the classifier and learns
    annotator-specific confusion matrices jointly with the model
    parameters via backpropagation \citep{Rodrigues_Pereira_2018}.
    \item \textbf{CoNAL:} Decomposes annotation noise into a common
    component shared across annotators and an individual component per
    annotator, using an auxiliary network to route each annotation to
    the appropriate noise source \citep{Chu_Ma_Wang_2021}.
    \item \textbf{AnnotMix:} Extends mixup-based training to the
    multi-annotator setting by interpolating both inputs and
    annotator-specific soft labels, regularising the model against
    annotation noise \citep{Herde2024}.
\end{itemize}

\section{Experimental Setup}
We largely follow \citet{Herde2024} for our experimental setup in the classifier experiments. 
Tab.~\ref{tab:dataset_splits} shows the data split for the classifier experiments. 
Tab.~\ref{tab:output_configs} shows the
hyperparameters used for the different baselines and \gls{IRT}-based methods.

\begin{table*}[t]
  \centering
  \caption{Data split statistics for all datasets. Val is a random 10\% of
  the crowd-annotated training data. The ground-truth-only validation splits
  published with Dopanim (750), MusicGenres (100) and SentimentPolarity (500),
  and the 80-item validation split of \gls{RTE}, are not used.}
  \label{tab:dataset_splits}
  \resizebox{0.6\textwidth}{!}{%
  \begin{tabular}{lrrrrrr}
    \toprule
    Dataset & Classes  & Train & Val & Test & Total \\
    \midrule
    Dopanim & 15 & 9,435 (63\%) & 1,049 (7\%) & 4,500 (30\%) & 14,984 \\
    MusicGenres & 10 & 630 (70\%) & 70 (8\%) & 200 (22\%) & 900 \\
    RTE & 2 & 504 (70\%) & 56 (8\%) & 160 (22\%) & 720 \\
    SentimentPolarity & 2 & 4,499 (45\%) & 500 (5\%) & 4,928 (50\%) & 9,927 \\
    CIFAR10H & 10 & 9,000 (15\%) & 1,000 (2\%) & 49,500 (83\%) & 59,500 \\
    CIFAR10N & 10 & 45,000 (76\%) & 5,000 (8\%) & 9,000 (15\%) & 59,000 \\
    \bottomrule
\end{tabular}}
\end{table*}

\begin{table*}[htbp]
    \centering
    \caption{Architecture and hyperparameter configurations per dataset and
		method. $K$ denotes the number of classes.
		\gls{IRT}/GLAD hyperparameters apply only to the \gls{IRT}/GLAD fitting with SVI.}
    \label{tab:output_configs}
    \resizebox{\textwidth}{!}{%
    \begin{tabular}{ll|c|c|c|c|c|c}
        \midrule
        \textbf{Method} & \textbf{Component}
            & \textbf{Dopanim}
            & \textbf{MusicGenres} & \textbf{SentimentPolarity}
            & \textbf{RTE} & \textbf{CIFAR10H}
            & \textbf{CIFAR10N} \\
        \midrule
        \multirow{7}{*}{All}
            & Backbone
                & DINOv2-base & MLP (256, 256, 256, 128)
                & MLP (256, 256, 256, 128) & DistilBERT-MNLI
                & ResNet-18 & ResNet-18 \\
            & Embed dim ($d$)
                & 768 & 128 & 128 & 768 & 512 & 512 \\
            & Output
                & $\mathrm{Lin}(d \to K)$ & $\mathrm{Lin}(d \to K)$
                & $\mathrm{Lin}(d \to K)$ & $\mathrm{Lin}(d \to K)$
                & $\mathrm{Lin}(d \to K)$ & $\mathrm{Lin}(d \to K)$ \\
            & Learning rate
                & $2 \times 10^{-5}$ & $10^{-2}$
                & $10^{-2}$ & $2 \times 10^{-5}$
                & $2 \times 10^{-5}$ & $2 \times 10^{-5}$ \\
            & Epochs
                & 20 & 100 & 100 & 20 & 20 & 20 \\
            & Batch size
                & 64 & 64 & 64 & 64 & 64 & 64 \\
            & Optimiser
                & RAdam & RAdam & RAdam & RAdam & RAdam & RAdam \\
        \midrule
        \multirow{2}{*}{IRT/GLAD}
            & SVI Learning rate
                & $0.02$ & $0.02$ & $0.02$ & $0.02$ & $0.02$ & $0.02$ \\
            & SVI Epochs
                & $10{,}000$ & $10{,}000$ & $10{,}000$
                & $10{,}000$ & $10{,}000$ & $10{,}000$ \\
        \midrule
        Crowd Layer
            & Conf.\ matrices $(K \!\times\! K \!\times\! R)$
                & $15 \!\times\! 15 \!\times\! 20$
                & $10 \!\times\! 10 \!\times\! 44$
                & $2 \!\times\! 2 \!\times\! 203$
                & $2 \!\times\! 2 \!\times\! 164$
                & $10 \!\times\! 10 \!\times\! 2{,}571$
                & $10 \!\times\! 10 \!\times\! 747$ \\
        \midrule
        \multirow{2}{*}{CoNAL}
            & Indiv.\ conf.\ $(R \!\times\! K \!\times\! K)$
                & $20 \!\times\! 15 \!\times\! 15$
                & $44 \!\times\! 10 \!\times\! 10$
                & $203 \!\times\! 2 \!\times\! 2$
                & $164 \!\times\! 2 \!\times\! 2$
                & $2{,}571 \!\times\! 10 \!\times\! 10$
                & $747 \!\times\! 10 \!\times\! 10$ \\
            & Common conf.\ $(K \!\times\! K)$
                & $15 \!\times\! 15$
                & $10 \!\times\! 10$
                & $2 \!\times\! 2$ & $2 \!\times\! 2$
                & $10 \!\times\! 10$ & $10 \!\times\! 10$ \\
        \midrule
        \multirow{2}{*}{AnnotMix}
            & AP hidden
                & \multicolumn{6}{c}{Dropout(0.5) $\to$ Lin($2h, h$) $\to$ BN($h$) $\to$ ReLU} \\
            & AP output $(h \to K^{\,2})$
                & $\mathrm{Lin}(128, 225)$
                & $\mathrm{Lin}(128, 100)$
                & $\mathrm{Lin}(128, 4)$
                & $\mathrm{Lin}(128, 4)$
                & $\mathrm{Lin}(128, 100)$ & $\mathrm{Lin}(128, 100)$ \\
        \midrule
    \end{tabular}}
\end{table*}

\section{Full Results}

The numbers behind the plots shown in the main body can be found in Tab.~\ref{tab:benchmark_results} (label-inference) and Tab.~\ref{tab:classifier_results} (learning from crowds experiments).

\begin{table*}[htbp]
	\centering
	\caption{Label inference results for the baselines and \gls{IRT} models on
	all datasets. All scores represent averages weighted by class.
	Bold indicates the best performance, and italics indicate the
	second-best.
	The values in brackets ($\Delta$lower, $\Delta$upper) are the bounds
	of the 95\% CI of the difference against hard~\gls{MV}, calculated
	using 1k paired bootstrap resamples.}
  	\label{tab:benchmark_results}
\resizebox{\linewidth}{!}{
  \begin{tabular}{lcccccc}
    \toprule
     & \multicolumn{3}{c}{\textbf{Dopanim}} & \multicolumn{3}{c}{\textbf{MusicGenres}} \\
    \cmidrule(lr){2-4} \cmidrule(lr){5-7}
    \textbf{Model} & \textbf{Brier Score} & \textbf{F1} & \textbf{ECE} & \textbf{Brier Score} & \textbf{F1} & \textbf{ECE} \\
    \midrule
    Hard MV & 0.3739 & 0.8115 & 0.0249 & 0.5968 & 0.6935 & 0.0597 \\
    Soft MV & 0.3028 ($\Delta$-0.08, $\Delta$-0.06) & 0.8115 ($\Delta$0.00, $\Delta$0.00) & 0.0167 ($\Delta$-0.01, $\Delta$-0.01) & 0.4747 ($\Delta$-0.17, $\Delta$-0.08) & 0.6935 ($\Delta$0.00, $\Delta$0.00) & 0.0492 ($\Delta$-0.01, $\Delta$-0.00) \\
    Dawid-Skene & \textbf{0.2596 ($\Delta$-0.13, $\Delta$-0.10)} & 0.8429 ($\Delta$0.03, $\Delta$0.04) & \textbf{0.0143 ($\Delta$-0.01, $\Delta$-0.01)} & 0.3919 ($\Delta$-0.26, $\Delta$-0.16) & 0.7714 ($\Delta$0.05, $\Delta$0.11) & 0.0381 ($\Delta$-0.03, $\Delta$-0.02) \\
    SVI-GLAD & 0.3091 ($\Delta$-0.07, $\Delta$-0.06) & 0.8236 ($\Delta$0.01, $\Delta$0.02) & 0.0190 ($\Delta$-0.01, $\Delta$-0.01) & 0.4330 ($\Delta$-0.20, $\Delta$-0.12) & 0.7875 ($\Delta$0.07, $\Delta$0.12) & 0.0336 ($\Delta$-0.03, $\Delta$-0.02) \\
    MACE & 0.3146 ($\Delta$-0.07, $\Delta$-0.05) & 0.8227 ($\Delta$0.01, $\Delta$0.02) & 0.0199 ($\Delta$-0.01, $\Delta$-0.00) & 0.3973 ($\Delta$-0.25, $\Delta$-0.15) & \textbf{0.7979 ($\Delta$0.08, $\Delta$0.13)} & 0.0396 ($\Delta$-0.02, $\Delta$-0.01) \\
    WAWA & 0.2813 ($\Delta$-0.10, $\Delta$-0.08) & 0.8230 ($\Delta$0.01, $\Delta$0.02) & 0.0174 ($\Delta$-0.01, $\Delta$-0.01) & 0.3756 ($\Delta$-0.27, $\Delta$-0.18) & \textit{0.7958 ($\Delta$0.08, $\Delta$0.13)} & 0.0427 ($\Delta$-0.02, $\Delta$-0.01) \\
    \cmidrule{1-7}
    IRT-1PL & 0.3024 ($\Delta$-0.08, $\Delta$-0.06) & 0.8246 ($\Delta$0.01, $\Delta$0.02) & 0.0186 ($\Delta$-0.01, $\Delta$-0.01) & 0.4003 ($\Delta$-0.24, $\Delta$-0.15) & 0.7956 ($\Delta$0.08, $\Delta$0.13) & 0.0400 ($\Delta$-0.02, $\Delta$-0.01) \\
    IRT-2PL & 0.3077 ($\Delta$-0.07, $\Delta$-0.06) & 0.8239 ($\Delta$0.01, $\Delta$0.02) & 0.0189 ($\Delta$-0.01, $\Delta$-0.01) & 0.3820 ($\Delta$-0.26, $\Delta$-0.17) & 0.7950 ($\Delta$0.08, $\Delta$0.13) & 0.0362 ($\Delta$-0.03, $\Delta$-0.02) \\
    IRT-3PL & 0.3110 ($\Delta$-0.07, $\Delta$-0.06) & 0.8241 ($\Delta$0.01, $\Delta$0.02) & 0.0193 ($\Delta$-0.01, $\Delta$-0.00) & \textit{0.3725 ($\Delta$-0.27, $\Delta$-0.18)} & 0.7944 ($\Delta$0.08, $\Delta$0.13) & 0.0341 ($\Delta$-0.03, $\Delta$-0.02) \\
    \cmidrule{1-7}
    CM-IRT-1PL & 0.2679 ($\Delta$-0.12, $\Delta$-0.09) & \textit{0.8455 ($\Delta$0.03, $\Delta$0.04)} & 0.0161 ($\Delta$-0.01, $\Delta$-0.01) & 0.3831 ($\Delta$-0.26, $\Delta$-0.17) & 0.7857 ($\Delta$0.07, $\Delta$0.12) & 0.0360 ($\Delta$-0.03, $\Delta$-0.02) \\
    CM-IRT-2PL & \textit{0.2599 ($\Delta$-0.12, $\Delta$-0.10)} & 0.8438 ($\Delta$0.03, $\Delta$0.04) & \textit{0.0151 ($\Delta$-0.01, $\Delta$-0.01)} & \textbf{0.3696 ($\Delta$-0.27, $\Delta$-0.18)} & 0.7937 ($\Delta$0.07, $\Delta$0.13) & \textbf{0.0319 ($\Delta$-0.03, $\Delta$-0.02)} \\
    CM-IRT-3PL & 0.2601 ($\Delta$-0.12, $\Delta$-0.10) & \textbf{0.8458 ($\Delta$0.03, $\Delta$0.04)} & \textit{0.0151 ($\Delta$-0.01, $\Delta$-0.01)} & 0.3758 ($\Delta$-0.27, $\Delta$-0.17) & 0.7891 ($\Delta$0.07, $\Delta$0.12) & \textit{0.0325 ($\Delta$-0.03, $\Delta$-0.02)} \\
  \end{tabular}
  }
  \medskip
  \resizebox{\linewidth}{!}{
  \begin{tabular}{lcccccc}
    \toprule
     & \multicolumn{3}{c}{\textbf{SentimentPolarity}} & \multicolumn{3}{c}{\textbf{RTE}} \\
    \cmidrule(lr){2-4} \cmidrule(lr){5-7}
    \textbf{Model} & \textbf{Brier Score} & \textbf{F1} & \textbf{ECE} & \textbf{Brier Score} & \textbf{F1} & \textbf{ECE} \\
    \midrule
    Hard MV & 0.1109 & 0.8891 & 0.1109 & 0.0833 & 0.9167 & 0.0833 \\
    Soft MV & 0.0940 ($\Delta$-0.02, $\Delta$-0.01) & 0.8891 ($\Delta$0.00, $\Delta$0.00) & 0.0794 ($\Delta$-0.05, $\Delta$-0.02) & 0.1034 ($\Delta$0.00, $\Delta$0.03) & 0.9167 ($\Delta$0.00, $\Delta$0.00) & 0.1671 ($\Delta$0.04, $\Delta$0.12) \\
    Dawid-Skene & 0.0755 ($\Delta$-0.04, $\Delta$-0.03) & 0.9151 ($\Delta$0.02, $\Delta$0.03) & 0.0660 ($\Delta$-0.05, $\Delta$-0.04) & 0.0661 ($\Delta$-0.03, $\Delta$-0.00) & \textit{0.9291 ($\Delta$-0.00, $\Delta$0.03)} & 0.0658 ($\Delta$-0.03, $\Delta$-0.00) \\
    SVI-GLAD & \textbf{0.0688 ($\Delta$-0.05, $\Delta$-0.04)} & \textit{0.9175 ($\Delta$0.02, $\Delta$0.03)} & \textbf{0.0480 ($\Delta$-0.07, $\Delta$-0.05)} & \textbf{0.0553 ($\Delta$-0.04, $\Delta$-0.01)} & 0.9264 ($\Delta$-0.01, $\Delta$0.03) & \textbf{0.0379 ($\Delta$-0.06, $\Delta$-0.02)} \\
    MACE & 0.0749 ($\Delta$-0.04, $\Delta$-0.03) & 0.9155 ($\Delta$0.02, $\Delta$0.03) & 0.0648 ($\Delta$-0.05, $\Delta$-0.04) & 0.0688 ($\Delta$-0.03, $\Delta$0.00) & 0.9221 ($\Delta$-0.01, $\Delta$0.02) & 0.0681 ($\Delta$-0.03, $\Delta$0.00) \\
    WAWA & 0.0862 ($\Delta$-0.03, $\Delta$-0.02) & 0.8944 ($\Delta$0.00, $\Delta$0.01) & 0.0722 ($\Delta$-0.05, $\Delta$-0.02) & 0.0837 ($\Delta$-0.01, $\Delta$0.01) & 0.9208 ($\Delta$0.00, $\Delta$0.01) & 0.1449 ($\Delta$0.03, $\Delta$0.09) \\
    \cmidrule{1-7}
    IRT-1PL & 0.0739 ($\Delta$-0.04, $\Delta$-0.03) & \textbf{0.9184 ($\Delta$0.02, $\Delta$0.04)} & 0.0639 ($\Delta$-0.05, $\Delta$-0.04) & 0.0733 ($\Delta$-0.03, $\Delta$0.01) & 0.9193 ($\Delta$-0.02, $\Delta$0.02) & 0.0742 ($\Delta$-0.03, $\Delta$0.01) \\
    IRT-2PL & \textit{0.0712 ($\Delta$-0.05, $\Delta$-0.03)} & 0.9166 ($\Delta$0.02, $\Delta$0.03) & \textit{0.0543 ($\Delta$-0.06, $\Delta$-0.05)} & 0.0645 ($\Delta$-0.03, $\Delta$-0.00) & 0.9263 ($\Delta$-0.01, $\Delta$0.03) & 0.0612 ($\Delta$-0.04, $\Delta$-0.00) \\
    IRT-3PL & 0.0722 ($\Delta$-0.04, $\Delta$-0.03) & 0.9160 ($\Delta$0.02, $\Delta$0.03) & 0.0556 ($\Delta$-0.06, $\Delta$-0.05) & \textit{0.0562 ($\Delta$-0.04, $\Delta$-0.01)} & \textbf{0.9305 ($\Delta$-0.00, $\Delta$0.03)} & \textit{0.0499 ($\Delta$-0.05, $\Delta$-0.02)} \\
    \cmidrule{1-7}
    CM-IRT-1PL & 0.0734 ($\Delta$-0.04, $\Delta$-0.03) & 0.9173 ($\Delta$0.02, $\Delta$0.04) & 0.0650 ($\Delta$-0.05, $\Delta$-0.04) & 0.0592 ($\Delta$-0.04, $\Delta$-0.01) & \textbf{0.9305 ($\Delta$-0.00, $\Delta$0.03)} & 0.0588 ($\Delta$-0.04, $\Delta$-0.01) \\
    CM-IRT-2PL & 0.0719 ($\Delta$-0.04, $\Delta$-0.03) & 0.9164 ($\Delta$0.02, $\Delta$0.03) & 0.0566 ($\Delta$-0.06, $\Delta$-0.05) & 0.0582 ($\Delta$-0.04, $\Delta$-0.01) & \textbf{0.9305 ($\Delta$-0.00, $\Delta$0.03)} & 0.0593 ($\Delta$-0.04, $\Delta$-0.01) \\
    CM-IRT-3PL & 0.0723 ($\Delta$-0.04, $\Delta$-0.03) & 0.9153 ($\Delta$0.02, $\Delta$0.03) & 0.0562 ($\Delta$-0.06, $\Delta$-0.05) & 0.0602 ($\Delta$-0.04, $\Delta$-0.01) & \textbf{0.9305 ($\Delta$-0.00, $\Delta$0.03)} & 0.0598 ($\Delta$-0.04, $\Delta$-0.01) \\
  \end{tabular}
  }
  \medskip
  \resizebox{\linewidth}{!}{
  \begin{tabular}{lcccccc}
    \toprule
     & \multicolumn{3}{c}{\textbf{CIFAR10H}} & \multicolumn{3}{c}{\textbf{CIFAR10N}} \\
    \cmidrule(lr){2-4} \cmidrule(lr){5-7}
    \textbf{Model} & \textbf{Brier Score} & \textbf{F1} & \textbf{ECE} & \textbf{Brier Score} & \textbf{F1} & \textbf{ECE} \\
    \midrule
    Hard MV & 0.0149 & 0.9925 & \textit{0.0015} & 0.1771 & 0.9116 & 0.0177 \\
    Soft MV & 0.0206 ($\Delta$0.00, $\Delta$0.01) & 0.9925 ($\Delta$0.00, $\Delta$0.00) & 0.0087 ($\Delta$0.01, $\Delta$0.01) & 0.1637 ($\Delta$-0.02, $\Delta$-0.01) & 0.9116 ($\Delta$0.00, $\Delta$0.00) & 0.0185 ($\Delta$0.00, $\Delta$0.00) \\
    Dawid-Skene & \textbf{0.0142 ($\Delta$-0.00, $\Delta$0.00)} & \textbf{0.9929 ($\Delta$-0.00, $\Delta$0.00)} & \textbf{0.0014 ($\Delta$-0.00, $\Delta$0.00)} & 0.1713 ($\Delta$-0.01, $\Delta$-0.00) & 0.9079 ($\Delta$-0.01, $\Delta$-0.00) & 0.0153 ($\Delta$-0.00, $\Delta$-0.00) \\
    SVI-GLAD & 0.0146 ($\Delta$-0.00, $\Delta$0.00) & 0.9924 ($\Delta$-0.00, $\Delta$0.00) & \textit{0.0015 ($\Delta$-0.00, $\Delta$0.00)} & \textbf{0.1313 ($\Delta$-0.05, $\Delta$-0.04)} & 0.9117 ($\Delta$-0.00, $\Delta$0.00) & \textbf{0.0071 ($\Delta$-0.01, $\Delta$-0.01)} \\
    MACE & 0.0149 ($\Delta$-0.00, $\Delta$0.00) & 0.9925 ($\Delta$-0.00, $\Delta$0.00) & \textit{0.0015 ($\Delta$-0.00, $\Delta$0.00)} & 0.1346 ($\Delta$-0.05, $\Delta$-0.04) & 0.9114 ($\Delta$-0.00, $\Delta$0.00) & 0.0098 ($\Delta$-0.01, $\Delta$-0.01) \\
    WAWA & 0.0200 ($\Delta$0.00, $\Delta$0.01) & 0.9925 ($\Delta$-0.00, $\Delta$0.00) & 0.0080 ($\Delta$0.01, $\Delta$0.01) & 0.1633 ($\Delta$-0.02, $\Delta$-0.01) & 0.9117 ($\Delta$-0.00, $\Delta$0.00) & 0.0184 ($\Delta$-0.00, $\Delta$0.00) \\
    \cmidrule{1-7}
    IRT-1PL & \textit{0.0144 ($\Delta$-0.00, $\Delta$0.00)} & \textit{0.9928 ($\Delta$-0.00, $\Delta$0.00)} & \textbf{0.0014 ($\Delta$-0.00, $\Delta$0.00)} & 0.1334 ($\Delta$-0.05, $\Delta$-0.04) & 0.9116 ($\Delta$-0.00, $\Delta$0.00) & 0.0092 ($\Delta$-0.01, $\Delta$-0.01) \\
    IRT-2PL & 0.0146 ($\Delta$-0.00, $\Delta$0.00) & 0.9925 ($\Delta$-0.00, $\Delta$0.00) & \textit{0.0015 ($\Delta$-0.00, $\Delta$0.00)} & 0.1322 ($\Delta$-0.05, $\Delta$-0.04) & 0.9119 ($\Delta$-0.00, $\Delta$0.00) & 0.0083 ($\Delta$-0.01, $\Delta$-0.01) \\
    IRT-3PL & 0.0147 ($\Delta$-0.00, $\Delta$0.00) & 0.9925 ($\Delta$-0.00, $\Delta$0.00) & \textit{0.0015 ($\Delta$-0.00, $\Delta$0.00)} & \textit{0.1321 ($\Delta$-0.05, $\Delta$-0.04)} & 0.9105 ($\Delta$-0.00, $\Delta$0.00) & \textit{0.0082 ($\Delta$-0.01, $\Delta$-0.01)} \\
    \cmidrule{1-7}
    CM-IRT-1PL & 0.0149 ($\Delta$-0.00, $\Delta$0.00) & 0.9925 ($\Delta$-0.00, $\Delta$0.00) & \textit{0.0015 ($\Delta$-0.00, $\Delta$0.00)} & 0.1346 ($\Delta$-0.05, $\Delta$-0.04) & \textbf{0.9151 ($\Delta$0.00, $\Delta$0.00)} & 0.0100 ($\Delta$-0.01, $\Delta$-0.01) \\
    CM-IRT-2PL & 0.0149 ($\Delta$-0.00, $\Delta$0.00) & 0.9925 ($\Delta$-0.00, $\Delta$0.00) & \textit{0.0015 ($\Delta$-0.00, $\Delta$0.00)} & 0.1332 ($\Delta$-0.05, $\Delta$-0.04) & \textbf{0.9151 ($\Delta$0.00, $\Delta$0.00)} & 0.0092 ($\Delta$-0.01, $\Delta$-0.01) \\
    CM-IRT-3PL & 0.0149 ($\Delta$-0.00, $\Delta$0.00) & 0.9925 ($\Delta$-0.00, $\Delta$0.00) & \textit{0.0015 ($\Delta$-0.00, $\Delta$0.00)} & 0.1332 ($\Delta$-0.05, $\Delta$-0.04) & \textit{0.9150 ($\Delta$0.00, $\Delta$0.00)} & 0.0092 ($\Delta$-0.01, $\Delta$-0.01) \\
    \bottomrule
\end{tabular}}
 \end{table*}

\begin{figure*}[t]
	\centering
	\includegraphics[width=0.83\linewidth]{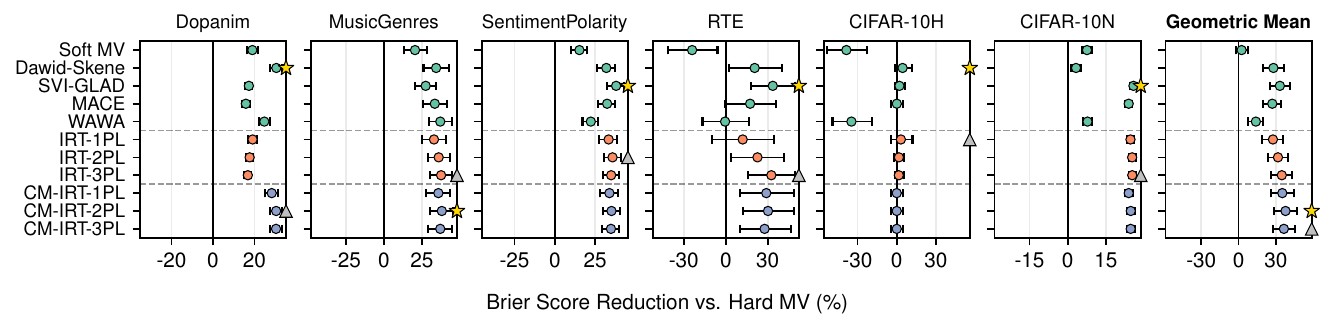}
	\caption{Label-inference benchmark Brier Score improvement relative
	to hard~\gls{MV} across all datasets.
	A gold star marks the best model per dataset and a silver triangle
	marks the second-best.
	Error bars indicate the 95\% CI calculated by 1k paired bootstrap
	resamples of the difference against hard~\gls{MV}.}
	\label{fig:benchmark_barplots_brier}
\end{figure*}

\begin{table*}[htbp]
	\centering
	\caption{Classification results for the baselines and \gls{IRT} models on
	all datasets. All scores are weighted by class prevalence.
	Bold indicates the best performance, and italics indicate the
	second-best.
	The $\pm$ values are the standard error of the mean across 10 random
	seeds.}
  	\label{tab:classifier_results}
  \resizebox{\linewidth}{!}{
  \begin{tabular}{lccccccccc}
    \toprule
     & \multicolumn{3}{c}{\textbf{Dopanim}} & \multicolumn{3}{c}{\textbf{MusicGenres}} & \multicolumn{3}{c}{\textbf{SentimentPolarity}} \\
    \cmidrule(lr){2-4} \cmidrule(lr){5-7} \cmidrule(lr){8-10}
    \textbf{Model} & \textbf{Brier Score} & \textbf{F1} & \textbf{ECE} & \textbf{Brier Score} & \textbf{F1} & \textbf{ECE} & \textbf{Brier Score} & \textbf{F1} & \textbf{ECE} \\
    \midrule
    Hard MV & 0.276 $\pm$ 0.01 & 0.798 $\pm$ 0.01 & 0.015 $\pm$ 0.00 & 0.471 $\pm$ 0.01 & 0.635 $\pm$ 0.01 & 0.047 $\pm$ 0.00 & 0.205 $\pm$ 0.00 & 0.688 $\pm$ 0.01 & \textit{0.042 $\pm$ 0.01} \\
    Soft MV & 0.316 $\pm$ 0.01 & 0.790 $\pm$ 0.01 & 0.024 $\pm$ 0.00 & 0.510 $\pm$ 0.01 & 0.663 $\pm$ 0.02 & 0.058 $\pm$ 0.00 & \textbf{0.195 $\pm$ 0.00} & 0.707 $\pm$ 0.01 & \textbf{0.040 $\pm$ 0.01} \\
    Dawid-Skene & 0.253 $\pm$ 0.01 & 0.819 $\pm$ 0.01 & \textit{0.013 $\pm$ 0.00} & 0.456 $\pm$ 0.01 & 0.677 $\pm$ 0.02 & \textbf{0.038 $\pm$ 0.00} & 0.200 $\pm$ 0.00 & 0.695 $\pm$ 0.00 & \textbf{0.040 $\pm$ 0.01} \\
    \cmidrule{1-10}
    Crowd Layer & 0.486 $\pm$ 0.05 & 0.708 $\pm$ 0.04 & 0.032 $\pm$ 0.00 & 0.576 $\pm$ 0.06 & 0.649 $\pm$ 0.04 & 0.060 $\pm$ 0.01 & 0.223 $\pm$ 0.00 & 0.705 $\pm$ 0.00 & 0.159 $\pm$ 0.01 \\
    CoNAL & 0.313 $\pm$ 0.01 & 0.805 $\pm$ 0.00 & 0.019 $\pm$ 0.00 & 0.497 $\pm$ 0.02 & 0.674 $\pm$ 0.02 & 0.051 $\pm$ 0.00 & 0.200 $\pm$ 0.00 & \textit{0.709 $\pm$ 0.00} & 0.073 $\pm$ 0.02 \\
    AnnotMix & 0.297 $\pm$ 0.01 & 0.814 $\pm$ 0.01 & 0.027 $\pm$ 0.00 & 0.425 $\pm$ 0.01 & \textbf{0.695 $\pm$ 0.01} & \textit{0.039 $\pm$ 0.00} & \textit{0.198 $\pm$ 0.00} & \textbf{0.714 $\pm$ 0.00} & 0.080 $\pm$ 0.01 \\
    \cmidrule{1-10}
    CM-IRT-1PL (Curriculum) & 0.257 $\pm$ 0.01 & 0.816 $\pm$ 0.01 & \textit{0.013 $\pm$ 0.00} & 0.428 $\pm$ 0.02 & \textit{0.691 $\pm$ 0.02} & \textit{0.039 $\pm$ 0.00} & 0.215 $\pm$ 0.00 & 0.675 $\pm$ 0.01 & 0.088 $\pm$ 0.02 \\
    CM-IRT-2PL (Curriculum) & \textit{0.248 $\pm$ 0.01} & \textit{0.824 $\pm$ 0.01} & \textbf{0.012 $\pm$ 0.00} & \textit{0.421 $\pm$ 0.01} & 0.688 $\pm$ 0.02 & \textit{0.039 $\pm$ 0.00} & 0.205 $\pm$ 0.00 & 0.691 $\pm$ 0.01 & 0.063 $\pm$ 0.01 \\
    CM-IRT-3PL (Curriculum) & \textbf{0.245 $\pm$ 0.01} & \textbf{0.825 $\pm$ 0.01} & \textbf{0.012 $\pm$ 0.00} & 0.434 $\pm$ 0.01 & 0.673 $\pm$ 0.02 & 0.040 $\pm$ 0.00 & 0.209 $\pm$ 0.00 & 0.683 $\pm$ 0.01 & 0.068 $\pm$ 0.01 \\
    \cmidrule{1-10}
    CM-IRT-1PL (Joint) & 0.330 $\pm$ 0.01 & 0.784 $\pm$ 0.01 & 0.024 $\pm$ 0.00 & \textbf{0.418 $\pm$ 0.02} & 0.683 $\pm$ 0.02 & 0.040 $\pm$ 0.00 & 0.207 $\pm$ 0.00 & 0.702 $\pm$ 0.00 & 0.093 $\pm$ 0.02 \\
    CM-IRT-2PL (Joint) & 0.333 $\pm$ 0.01 & 0.777 $\pm$ 0.02 & 0.025 $\pm$ 0.00 & 0.437 $\pm$ 0.02 & 0.679 $\pm$ 0.02 & 0.041 $\pm$ 0.00 & 0.206 $\pm$ 0.00 & 0.701 $\pm$ 0.00 & 0.089 $\pm$ 0.01 \\
    CM-IRT-3PL (Joint) & 0.332 $\pm$ 0.01 & 0.779 $\pm$ 0.01 & 0.025 $\pm$ 0.00 & 0.426 $\pm$ 0.01 & 0.684 $\pm$ 0.02 & 0.041 $\pm$ 0.00 & 0.204 $\pm$ 0.00 & 0.702 $\pm$ 0.01 & 0.077 $\pm$ 0.02 \\
    \cmidrule{1-10}
    Oracle & 0.132 $\pm$ 0.01 & 0.909 $\pm$ 0.00 & 0.006 $\pm$ 0.00 & 0.332 $\pm$ 0.01 & 0.783 $\pm$ 0.01 & 0.032 $\pm$ 0.00 & 0.199 $\pm$ 0.00 & 0.697 $\pm$ 0.01 & 0.035 $\pm$ 0.01 \\
  \end{tabular}
  }
  \medskip
  \resizebox{\linewidth}{!}{
  \begin{tabular}{lccccccccc}
    \toprule
     & \multicolumn{3}{c}{\textbf{RTE}} & \multicolumn{3}{c}{\textbf{CIFAR10H}} & \multicolumn{3}{c}{\textbf{CIFAR10N}} \\
    \cmidrule(lr){2-4} \cmidrule(lr){5-7} \cmidrule(lr){8-10}
    \textbf{Model} & \textbf{Brier Score} & \textbf{F1} & \textbf{ECE} & \textbf{Brier Score} & \textbf{F1} & \textbf{ECE} & \textbf{Brier Score} & \textbf{F1} & \textbf{ECE} \\
    \midrule
    Hard MV & 0.212 $\pm$ 0.01 & 0.682 $\pm$ 0.02 & 0.106 $\pm$ 0.01 & 0.170 $\pm$ 0.00 & 0.885 $\pm$ 0.00 & 0.009 $\pm$ 0.00 & 0.132 $\pm$ 0.00 & 0.914 $\pm$ 0.00 & 0.013 $\pm$ 0.00 \\
    Soft MV & 0.208 $\pm$ 0.01 & \textit{0.691 $\pm$ 0.03} & \textbf{0.092 $\pm$ 0.01} & 0.136 $\pm$ 0.00 & 0.909 $\pm$ 0.00 & 0.009 $\pm$ 0.00 & 0.156 $\pm$ 0.00 & 0.921 $\pm$ 0.00 & 0.032 $\pm$ 0.00 \\
    Dawid-Skene & 0.213 $\pm$ 0.01 & 0.680 $\pm$ 0.02 & 0.104 $\pm$ 0.02 & 0.173 $\pm$ 0.00 & 0.884 $\pm$ 0.00 & 0.010 $\pm$ 0.00 & 0.153 $\pm$ 0.01 & 0.905 $\pm$ 0.00 & 0.020 $\pm$ 0.00 \\
    \cmidrule{1-10}
    Crowd Layer & 0.223 $\pm$ 0.01 & 0.687 $\pm$ 0.02 & 0.163 $\pm$ 0.02 & 0.246 $\pm$ 0.06 & 0.837 $\pm$ 0.05 & 0.023 $\pm$ 0.01 & 0.141 $\pm$ 0.05 & 0.917 $\pm$ 0.03 & 0.013 $\pm$ 0.00 \\
    CoNAL & \textit{0.207 $\pm$ 0.01} & \textbf{0.694 $\pm$ 0.03} & 0.101 $\pm$ 0.01 & 0.163 $\pm$ 0.00 & 0.902 $\pm$ 0.00 & 0.014 $\pm$ 0.00 & \textbf{0.120 $\pm$ 0.00} & \textit{0.927 $\pm$ 0.00} & \textbf{0.010 $\pm$ 0.00} \\
    AnnotMix & \textbf{0.206 $\pm$ 0.01} & \textit{0.691 $\pm$ 0.03} & \textit{0.098 $\pm$ 0.01} & 0.130 $\pm$ 0.00 & \textit{0.918 $\pm$ 0.00} & 0.015 $\pm$ 0.00 & 0.124 $\pm$ 0.00 & \textbf{0.931 $\pm$ 0.00} & 0.023 $\pm$ 0.00 \\
    \cmidrule{1-10}
    CM-IRT-1PL (Curriculum) & 0.211 $\pm$ 0.01 & 0.687 $\pm$ 0.01 & 0.101 $\pm$ 0.02 & 0.147 $\pm$ 0.00 & 0.901 $\pm$ 0.00 & \textbf{0.005 $\pm$ 0.00} & 0.128 $\pm$ 0.00 & 0.918 $\pm$ 0.00 & 0.013 $\pm$ 0.00 \\
    CM-IRT-2PL (Curriculum) & 0.211 $\pm$ 0.01 & 0.686 $\pm$ 0.02 & 0.102 $\pm$ 0.02 & 0.234 $\pm$ 0.00 & 0.838 $\pm$ 0.00 & 0.009 $\pm$ 0.00 & 0.128 $\pm$ 0.00 & 0.918 $\pm$ 0.00 & 0.013 $\pm$ 0.00 \\
    CM-IRT-3PL (Curriculum) & 0.217 $\pm$ 0.01 & 0.681 $\pm$ 0.01 & 0.113 $\pm$ 0.04 & 0.230 $\pm$ 0.01 & 0.842 $\pm$ 0.00 & 0.009 $\pm$ 0.00 & 0.128 $\pm$ 0.00 & 0.918 $\pm$ 0.00 & 0.013 $\pm$ 0.00 \\
    \cmidrule{1-10}
    CM-IRT-1PL (Joint) & 0.246 $\pm$ 0.02 & 0.680 $\pm$ 0.01 & 0.200 $\pm$ 0.05 & 0.121 $\pm$ 0.00 & \textit{0.918 $\pm$ 0.00} & \textit{0.006 $\pm$ 0.00} & \textit{0.122 $\pm$ 0.00} & 0.920 $\pm$ 0.00 & \textit{0.012 $\pm$ 0.00} \\
    CM-IRT-2PL (Joint) & 0.213 $\pm$ 0.01 & 0.677 $\pm$ 0.02 & 0.106 $\pm$ 0.03 & \textit{0.117 $\pm$ 0.00} & \textbf{0.921 $\pm$ 0.00} & \textit{0.006 $\pm$ 0.00} & 0.138 $\pm$ 0.01 & 0.920 $\pm$ 0.00 & 0.022 $\pm$ 0.01 \\
    CM-IRT-3PL (Joint) & 0.211 $\pm$ 0.01 & 0.673 $\pm$ 0.02 & \textit{0.098 $\pm$ 0.02} & \textbf{0.116 $\pm$ 0.00} & \textbf{0.921 $\pm$ 0.00} & \textbf{0.005 $\pm$ 0.00} & 0.129 $\pm$ 0.01 & 0.918 $\pm$ 0.00 & 0.015 $\pm$ 0.01 \\
    \cmidrule{1-10}
    Oracle & 0.210 $\pm$ 0.01 & 0.677 $\pm$ 0.03 & 0.088 $\pm$ 0.02 & 0.166 $\pm$ 0.00 & 0.889 $\pm$ 0.00 & 0.009 $\pm$ 0.00 & 0.101 $\pm$ 0.00 & 0.933 $\pm$ 0.00 & 0.006 $\pm$ 0.00 \\
    \bottomrule
  \end{tabular}
  }\end{table*}

\begin{figure*}[!t]
	\centering
	\subfloat[\label{fig:classifier_barplots_brier}]{%
		\includegraphics[width=0.93\linewidth]{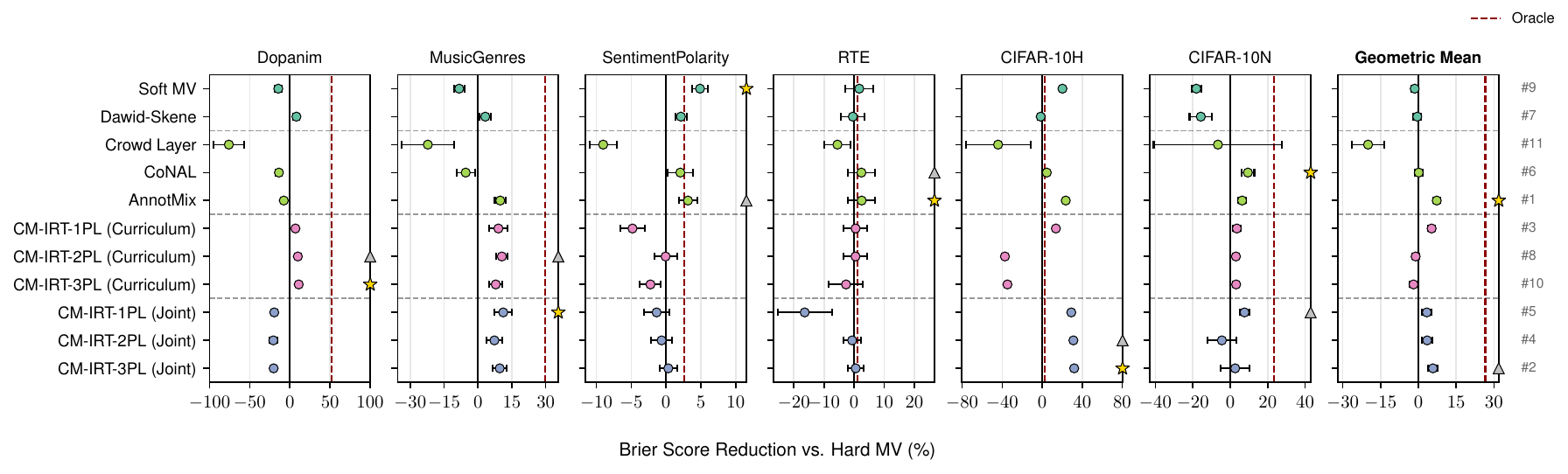}}\\
	\subfloat[\label{fig:classifier_barplots_f1}]{%
		\includegraphics[width=0.93\linewidth]{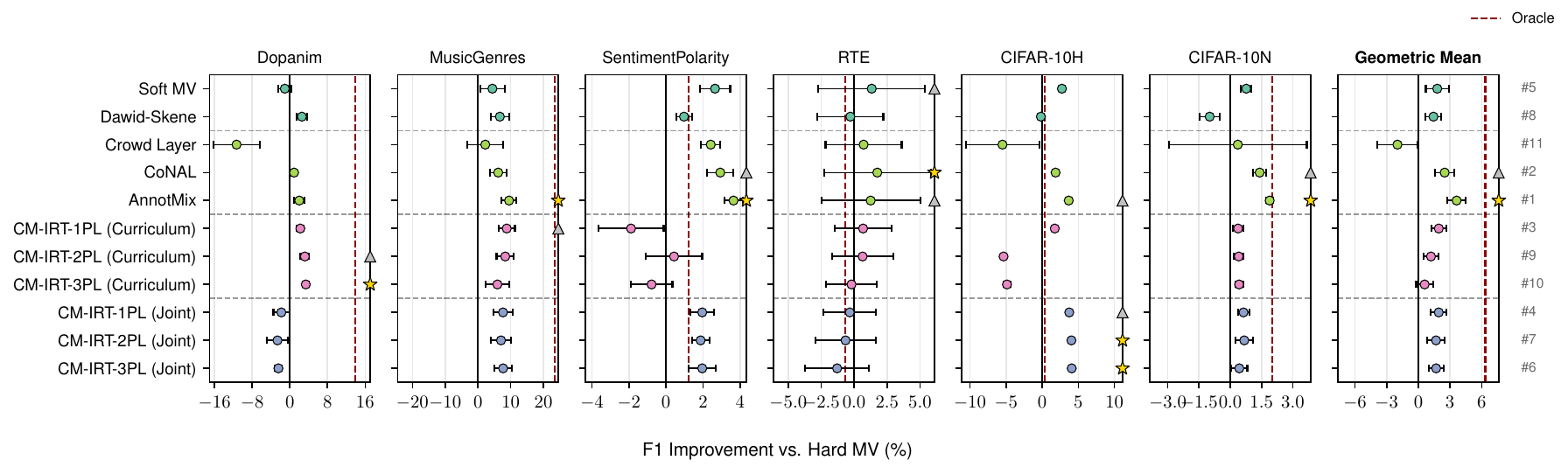}}
	\caption{Classifier results across all datasets.
	(a)~Brier Score improvement and (b)~F1 improvement,
	both relative to hard~\gls{MV}.
	A gold star marks the best model per dataset and
	a silver triangle marks the second-best.
	The complete results can be found in Tab.~\ref{tab:classifier_results}.
	Significance against hard~\gls{MV} is assessed via a seed-paired
	permutation test (App.~\ref{app:sec:perm_test}).}
	\label{fig:classifier_barplots}
\end{figure*}

\section{Significance Testing}
\label{app:sec:perm_test}

We compare each method against hard~\gls{MV} with a seed-paired permutation test.
We pair each run of a method with the hard~\gls{MV} run on the same seed.
We summarise each pair by the log-ratio of the metric over the baseline (F1) or
inverted for losses (Brier Score, ECE).
The test statistic ($S_{obs}$) averages the log-ratios first within and then across
datasets,
which weights every dataset equally.
Under the null hypothesis that the method and hard~\gls{MV} are exchangeable, each
log-ratio is symmetric around zero. We therefore draw 10k random sign
assignments, flipping signs independently within each
dataset, and recompute the test statistic ($S_{perm}$) for each. The two-sided $p$-value is the fraction
of draws with $|S_{perm}| \geq |S_{obs}|$, with a $+1$ correction in
numerator and denominator.
We control the family-wise error rate across methods for each metric with the
Holm correction \citep{holm} and call a difference significant if the adjusted $p < 0.05$.
Confidence intervals are estimated using a Student's $t$ distribution and the standard error using the
within-dataset variance.

\section{Additional Experiments}
\label{app:sec:adv_raters}

\begin{figure*}[p]
	\centering
	\subfloat[\label{fig:adv_sweep}]{%
		\includegraphics[width=0.249\linewidth]{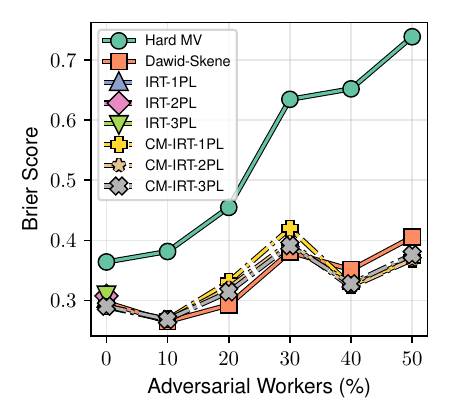}}
	\subfloat[\label{fig:adv_sweep_f1}]{%
		\includegraphics[width=0.249\linewidth]{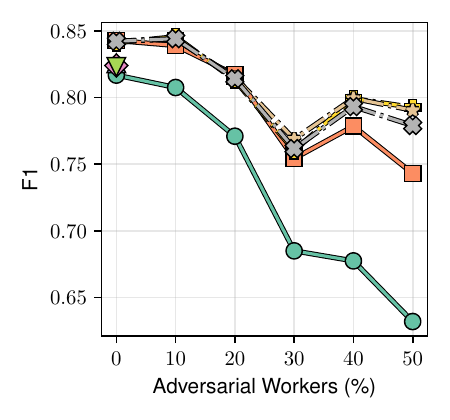}}
	\subfloat[\label{fig:raters_sweep}]{%
		\includegraphics[width=0.249\linewidth]{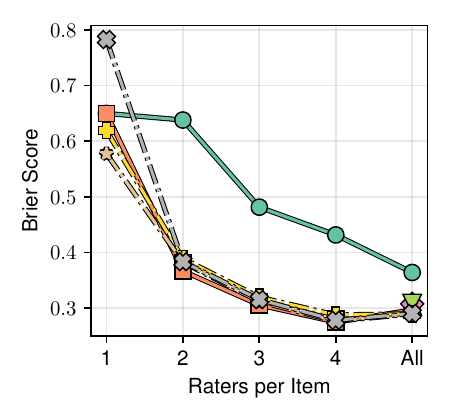}}
	\subfloat[\label{fig:raters_sweep_f1}]{%
		\includegraphics[width=0.249\linewidth]{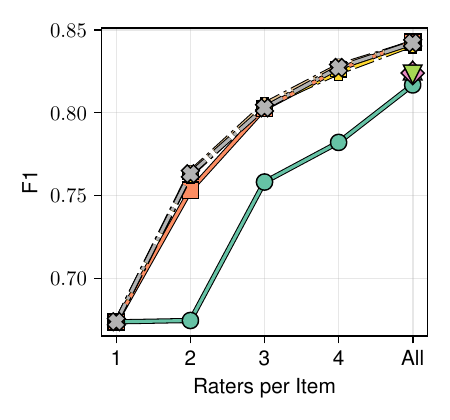}}
	\caption{(a)--(b)~CM-\gls{IRT} model performance under adversarial
	annotators on Dopanim, where 0--50\% of annotators' labels were
	randomised (flipped for binary, uniform random for multi-class)
	before recalculating the metrics, shown as Brier Score and F1
	respectively.
	(c)--(d)~Uniform \gls{IRT} model performance on Dopanim as a function of
	the number of raters per item, shown as Brier Score and F1
	respectively.}
	\label{fig:adv_and_raters}
\end{figure*}

\paragraph{Adversarial Annotators.}
To test aggregation performance under adversarial annotators, we
conduct an experiment where we randomise all labels for 0--50\% of
annotators on Dopanim (flipping for binary, drawing uniformly at
random for multi-class) and rerun the benchmark.
Fig.~\ref{fig:adv_sweep} and~\ref{fig:adv_sweep_f1} show the
resulting Brier Score and F1 for \gls{IRT} models, the best baseline, and
majority vote.
All models are robust to a small number of adversaries, with the
utility of models against the majority vote increasing with the
number of adversaries.
Generally, \gls{IRT} models underperform Dawid-Skene with few adversarial
annotators and outperform it with more.

\paragraph{Number of Raters.}
To investigate how model performance scales with annotation density,
we evaluate all methods on Dopanim using subsets of 1, 2, 3, and 4
raters per item, in addition to the full annotation set.
Fig.~\ref{fig:raters_sweep} and~\ref{fig:raters_sweep_f1} show Brier
Score and F1 as a function of the number of raters per item.
We find that \gls{IRT} models outperform the
majority vote consistently over different numbers of raters.

\paragraph{Synthetic Crowd-Learning Experiments.}
\label{app:sec:synthetic_testbed}

\begin{figure*}[p]
	\centering
	\includegraphics[width=0.50\linewidth]{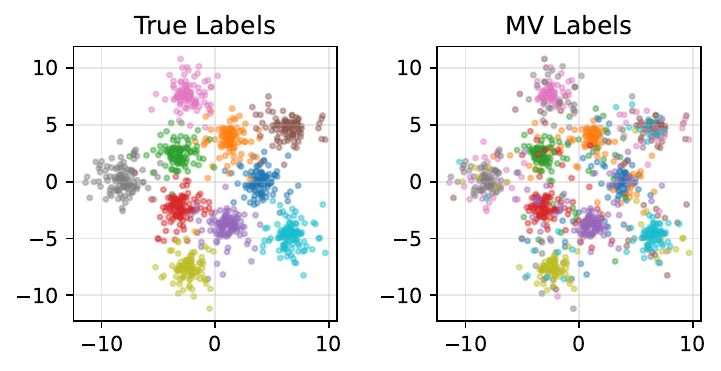}
	\caption{Visualisation of the synthetic testbed data. Clusters (classes) are not separable linearly and spread is dependent on item difficulty.
	The left-hand side shows clusters coloured with the real ground truth and the right-hand side shows them coloured by the \gls{MV}-label.}
	\label{fig:testbed_data_scatter}
\end{figure*}

To test whether our classifier works under perfectly fulfilled assumptions, we
conduct controlled synthetic experiments by generating annotations from a known
\gls{IRT}-2PL model.
The data consists of 10 perfectly separable blobs in 2D space, where each blob
represents a class.
The classes are not linearly separable due to the arrangement of the blobs in space
(Fig.~\ref{fig:testbed_data_scatter}).
We initially generate 1000 samples per class and then generate noisy annotations
for each sample by sampling from the \gls{IRT} model with known parameters.
For a crowd-learning approach to be beneficial, the majority-vote labels
must yield suboptimal performance on the downstream classifier, which we ensure by
setting the mean annotator ability ($\cann{\theta}_\text{pop} = 0.1$) close to the mean
item difficulty, resulting in approximately $45\%$ of individual annotations being
incorrect on average.
We then train various classifiers and compare them to baselines trained on soft
majority-vote labels (floor) and ground-truth labels (ceiling).
We assume that the difficulty of items is not independent of the features of the
items, i.e.\ we expect that items that are more difficult to label are also
more ambiguous in their features.
To simulate this, we increase the variance of the blobs in feature space for items
with higher difficulty.

\begin{figure*}[p]
	\centering
	\includegraphics[width=1.00\linewidth]{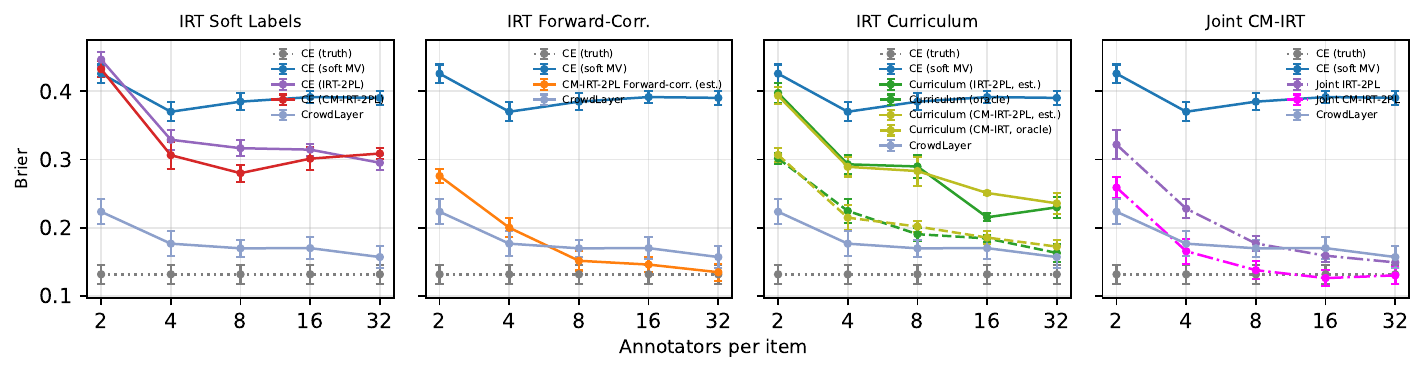}
	\caption{Brier Score of the downstream classifier as a function of the number of
	annotators per item in the synthetic experiment. Solid lines show 2PL variants,
	dashed lines show oracle classifiers which know the true \gls{IRT} parameters
	during fit.
	As a reference, we plot classifiers using soft majority-vote and ground-truth,
	plus the best performing crowd-method (Crowd Layer).
	Error bars indicate the standard error of the mean across 10-fold
	cross-validation.}
	\label{fig:testbed_final_brier}
\end{figure*}

\paragraph{Results.}
We find that under perfect assumptions, \gls{IRT}-informed approaches outperform
the best crowd-method as expected (Fig.~\ref{fig:testbed_final_brier}).
Gains are only visible with enough annotations per example, presumably
due to the challenge of estimating the discrimination parameter correctly under
data scarcity.
Notably, the joint model outperforms the crowd-baseline the fastest given more
annotators per item.

\end{document}